\documentclass[]{artificer}

\usepackage[utf8]{inputenc} 
\usepackage{hyperref}       
\usepackage{url}            
\usepackage{booktabs}       
\usepackage{amsfonts}       
\usepackage{nicefrac}       
\usepackage{microtype}      
\usepackage{xcolor}         
\usepackage{microtype}      
\usepackage{xcolor}         
\usepackage{amsmath}
\usepackage{graphicx}
\usepackage{nicefrac}       
\usepackage{amsfonts}       
\usepackage[utf8]{inputenc} 
\usepackage{booktabs}       
\usepackage{circledsteps}
\usepackage{multirow}
\usepackage{pifont}                         
\usepackage[dvipsnames]{xcolor}
\usepackage{circledsteps}
\usepackage{pifont}
\usepackage{wrapfig}
\usepackage{algorithm}
\usepackage{algpseudocode}
\usepackage{tikz}
\usepackage{tcolorbox}
\usepackage{dsfont}
\usepackage{enumitem}
\usepackage{marvosym}

\tcbuselibrary{breakable}
\newcommand{\gmark}{\textcolor{ForestGreen}{\ding{51}}}  
\newcommand{\xmark}{\textcolor{BrickRed}{\ding{55}}}
\usepackage{colortbl}  
\definecolor{ourcol}{HTML}{EAF1FB}    
\definecolor{headcol}{HTML}{E6E6E6}   

\definecolor{GainBright}{RGB}{0,114,178}   
\newcommand{\conf}[2]{#1{\scriptsize\,(\textcolor{GainBright}{#2})}}

\title{MILES: Modular Instruction Memory with Learnable Selection for Self-Improving LLM Reasoning}

\author[]{Ruilin Tong}
\author[{\textnormal{\Letter}}]{Dong Gong}
\affiliation[]{University of New South Wales (UNSW Sydney)}
\emails{\{ruilin.tong, dong.gong\}@unsw.edu.au}
\contribution{\textsuperscript{\Letter}Corresponding author.}

\abstract{Large language models (LLMs) increasingly improve their reasoning at test time via additional computation, yet most existing works treat each problem in isolation. When problems arrive sequentially, accumulating reusable experience across them can further improve performance. Existing memory-based methods either store whole-solution templates that generalize poorly to novel problems or use heuristic step-level selection that is not optimized for final-answer correctness. Learning selection policies requires large-scale training data and fixed action spaces, making such approaches unsuitable for test-time settings where memory expands incrementally and only limited supervision is available.
We propose \textbf{MILES} (\textbf{M}odular \textbf{I}nstruction Memory with \textbf{LE}arnable \textbf{S}election), a framework that dynamically expands step-wise memory and applies correctness-optimized memory composition under realistic test-time constraints.
MILES maintains modular memory units consisting of asymmetric pairs of sub-goal embeddings and sub-instructions, each associated with a learnable selection head. This memory structure enables a coarse-to-fine retrieval mechanism: The coarse level enables memory expansion and collects supervision for training selection heads from confident samples, while the fine stage applies learned selection heads to rerank coarse-level candidates and guide reasoning for uncertain samples. MILES achieves self-improvement by using its own confidence as the learning signal at test time, or given feedback such as labeled data in an offline setting.
MILES consistently matches or outperforms prior methods while achieving superior accuracy–efficiency tradeoffs. Extensive experiments demonstrate its effectiveness, robustness, and transferability.}

\metadata[Project Page]{\url{https://artificer-ai-lab.github.io/MILES/}}

\begin{document}

\maketitle

\section{Introduction}

Large language models (LLMs) demonstrate strong capabilities in complex reasoning and question answering. Recent advances in test-time computation, including chain-of-thought prompting~\cite{wei2022chain}, self-consistency~\cite{wang2022self}, and tree search~\cite{yao2023tree,qi2024mutual}, have further improved reasoning performance. In many practical settings, however, problems arrive sequentially rather than in isolation, allowing models to benefit from \emph{accumulating reusable experience} across related queries~\cite{yang2024buffer,ouyang2025reasoningbank,lu2026ascent}. Updating model parameters for every new problem is often costly or infeasible, especially for closed or frozen models~\cite{zelikman2024star,zeng2024b}. A natural alternative is therefore to maintain an \emph{external memory} that the LLM consults at test time while its parameters remain frozen~\cite{suzgun2025dynamic,zhang2025agentic,shinn2023reflexion}. The central question becomes: \emph{how can such a memory be organized, maintained, and reused so that it actually improves reasoning---and continues to improve as more experience accumulates?}

Existing memory methods for LLM reasoning store \emph{whole-solution} templates or strategies and retrieve them per problem~\cite{yang2024buffer,suzgun2025dynamic,zhang2025agentic}, which are useful when the new problem closely resembles a seen one, but limited when the new problem has novel structure. \emph{Step-level} memory~\cite{ahmed2025retrieval,stein2025experience,ho2025arcmemo} stores smaller units that can be composed flexibly across reasoning steps and across problems with different overall structures, shifting the problem from finding a similar past problem to finding applicable local blocks for the next reasoning step. Among step-level designs, prior work commits to all-text concept/situation pairs~\cite{ho2025arcmemo}, pre-formed templates~\cite{ahmed2025retrieval}, or free-form strategies~\cite{shinn2023reflexion,ouyang2025reasoningbank}. A common limitation persists across all of these: \emph{selecting} which unit to apply at the current step is heuristic, via similarity retrieval or LLM prompting, without optimizing the decision against final-answer correctness. Methods that learn selection from feedback~\cite{zhao2025dynaact,brooks2022context,zhou2025memento} require fixed action spaces, external labeled corpora, or fine-tuning, which are not compatible with a memory that grows incrementally from the model's own experience under limited test-time data. Moreover, policies learned offline may not generalize well as test data distribution shifts.

We propose \textbf{MILES} (\textbf{M}odular \textbf{I}nstruction Memory with \textbf{LE}arnable \textbf{S}election), a framework that enables test-time self-improvement by learning to compose reusable step-level reasoning modules from accumulated reasoning experience while keeping the underlying LLM frozen. To support reusable step-level reasoning, MILES stores memory as modular asymmetric pairs of a \emph{sub-goal} embedding and a corresponding \emph{sub-instruction}. The sub-goal embedding serves as a retrieval key representing the objective of the next reasoning step, while the sub-instruction is natural-language guidance used to condition generation. Unlike raw CoT traces, MILES stores reusable local reasoning modules. Its asymmetric (embedding, text) representation supports scalable retrieval and incremental memory expansion while preserving flexible instruction generation.
To enable \emph{learnable memory selection}, MILES associates each memory unit with a lightweight selection head and employs a coarse-to-fine retrieval mechanism at inference. A cheap sub-goal similarity-based retrieval stage (Layer~1) followed by selection-head applicability scoring (Layer~2); Layer~1 expands memory and collects dense step-level supervision from confident trajectories, enabling the selection heads to learn reusable memory-selection patterns from reliable reasoning experience. Layer~2 then applies the learned selection heads to guide reasoning on uncertain samples leveraging memory-selection patterns learned from confident trajectories. As a result, MILES turns step-wise memory composition into a learned, correctness-optimized decision under realistic test-time constraints: incremental memory growth, sparse per-unit data, no ground-truth labels, and no parameter updates. This enables MILES to self-improve at test time, using its own confidence (agreement among sampled answers) as the learning signal. In an offline setting, it can instead learn from given feedback such as labeled data (Table~\ref{tab:train-tranfer}). MILES consistently matches or outperforms prior methods across six reasoning benchmarks while achieving superior accuracy–efficiency tradeoffs. Extensive experiments further validate its effectiveness, robustness, and transferability. MILES, with its step-wise modular memory composition, complements prompting and test-time search by reusing experience across problems.


The main contributions of this work are summarized as follows: \vspace{-5pt}

\begin{itemize}[leftmargin=10pt, rightmargin=0pt]
    \item \textbf{The MILES framework} for self-improving step-wise memory composition at test time. MILES incrementally builds modular instruction memory from confident reasoning traces and applies the learned memory to uncertain cases, without parameter updates, external supervision, curated training data, or fixed action spaces.

    \item \textbf{A modular asymmetric memory structure} consisting of (sub-goal, sub-instruction) pairs associated with per-unit selection heads, naturally enabling a coarse-to-fine selection mechanism that enables both incremental memory expansion and learnable memory composition.

    \item \textbf{Learnable memory composition.} MILES trains lightweight per-unit selection heads from rollout outcomes on confident samples, enabling self-supervised and correctness-optimized memory composition under limited test-time data while supporting incremental memory expansion.

    \item \textbf{Empirical validation.} MILES consistently matches or outperforms baseline methods across six reasoning benchmarks and four backbones while achieving superior accuracy–efficiency tradeoffs. Further analyses demonstrate its effectiveness, robustness, and transferability.
\end{itemize}

\vspace{-5pt}
\section{Related work}
\vspace{-5pt}

\textbf{Test-time LLM reasoning.} Chain-of-Thought (CoT) prompting \cite{wei2022chain} and Self-Consistency (SC) \cite{wang2022self} improve LLM reasoning through intermediate reasoning and multi-trajectory aggregation. Structured-search methods such as Tree-of-Thought (ToT) \cite{yao2023tree}, Graph-of-Thought (GoT) \cite{besta2024graph}, rStar \cite{qi2024mutual}, Tree-BoN \cite{qiu2024treebon}, and DORA \cite{wang2025every} further improve reasoning through structured exploration and adaptive search. However, these methods search over memoryless rollouts without reusable cross-problem knowledge. In contrast, MILES integrates learned step-level memory into the search process through per-step sub-instruction selection.

\textbf{Improve LLM reasoning with memory.} Several recent works improve LLM reasoning with reusable memory derived from past experience. Buffer-of-Thought \cite{yang2024buffer}, Dynamic CheatSheet \cite{suzgun2025dynamic}, and ACE \cite{zhang2025agentic} maintain and retrieve problem-solving strategies during test time. A position paper~\cite{dorovatas2026positionmodularmemorykey} argues that modular memory is key to developing continually adaptable agents. Modular parametric memory designs are used to reduce forgetting in continual learning~\cite{lu2026little,zhao2026token}. ArcMemo \cite{ho2025arcmemo} and Retrieval-of-Thought \cite{ahmed2025retrieval} introduce step-level or template-based \cite{yang2025reasonflux} reasoning memory. However, these methods rely on heuristic memory selection via similarity or prompting rather than optimizing selection for final-answer correctness. Beyond reusable memory construction, recent works improve reasoning by updating memory or retrieval mechanisms using feedback. ICPI \cite{brooks2022context}, Memento \cite{zhou2025memento}, and DynaAct \cite{zhao2025dynaact} learn action utilities or value estimates from past trajectories, while ReasoningBank \cite{ouyang2025reasoningbank}, EGUR \cite{stein2025experience}, and BCI \cite{didolkar2025metacognitive} iteratively refine reusable strategies through self-feedback. ASCENT~\cite{lu2026ascent} trains agents online at test time by distilling verified experience into fast weights. These methods typically rely on fixed action spaces, external labeled data, or parameter updates, making them not compatible with incrementally growing, \emph{frozen-LLM} test-time memory. In contrast, MILES incrementally expands its memory while learning reusable memory-selection patterns from limited test-time data.

\vspace{-5pt}
\section{The Proposed Method: MILES}
\vspace{-5pt}
\label{sec:method}

\subsection{Problem formulation and setting}
\vspace{-5pt}
\label{sec:formulation}

\textbf{Test-time LLM reasoning.} In many practical settings, problems arrive sequentially. We consider a pretrained large language model (LLM) that solves a question $q$ by generating a reasoning trajectory $e_1, e_2, \ldots, e_T$ followed by a final answer $y$. We define the reasoning state after $t$ steps as $x_t=(q,e_1,\ldots,e_t)$ with $x_0=q$, and denote the step-level generation distribution by $p(e\mid x)$. Let $y^*$ denote the ground-truth answer and $r\in\{0,1\}$ the binary correctness indicator, where $r=1$ if $y=y^*$.

\textbf{Test-time improvement via external memory.} In a stream of test questions, accumulating reusable problem-solving experience into an \emph{external memory} $\mathcal{M}$ allows past experience to benefit subsequent reasoning. The LLM parameters remain frozen throughout, so all learning in MILES is performed by lightweight auxiliary modules built on top of the fixed generation model $p(\cdot \mid \cdot)$. 

\vspace{-5pt}
\subsection{Overview of MILES}
\vspace{-5pt}
\label{sec:overview}

\begin{figure}
    \centering
    \includegraphics[width=0.95\linewidth]{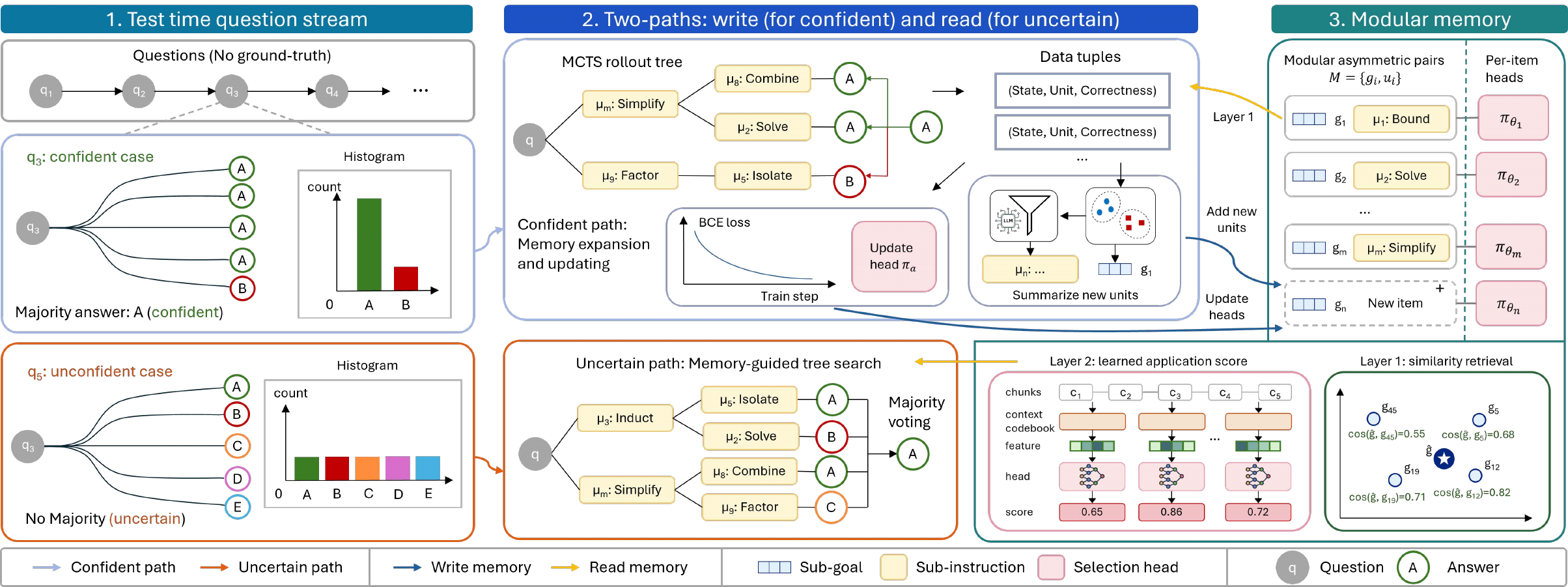}\vspace{-2pt}
    \caption{Overview of MILES. MILES improves test-time reasoning by constructing memory and learning memory selection from confident samples, then applying the learned memory to guide reasoning on uncertain samples.} \vspace{-10pt}
    \label{fig:method-overview}
\end{figure}

Our goal is to improve test-time reasoning by augmenting a frozen LLM with an external step-wise memory built from the model’s own reasoning experience. MILES maintains a memory $\mathcal{M}$ of modular (sub-goal embedding, sub-instruction text) units with lightweight per-unit selection heads. As the model processes a stream of test problems, MILES operates in two modes: (1) For confident samples, it incrementally expands memory and learns selection heads from rollout outcomes, enabling the system to accumulate reusable reasoning experience over time. (2) For uncertain samples, it retrieves and reranks relevant memory units through a coarse-to-fine selection mechanism to guide reasoning and improve accuracy. Figure~\ref{fig:method-overview} illustrates the overall framework of MILES. The remainder of this section details the four core components that realize this design.

We first define and construct modular memory units in Section~\ref{sec:mem-unit}. Building on this memory structure, Section~\ref{sec:mem-rs} introduces a coarse-to-fine selection mechanism that supports both memory expansion and training-data collection for the selection heads. Based on the collected supervision, Section~\ref{sec:learn-head} presents the learning procedure for the lightweight selection heads under limited data and incrementally expanding memory. Finally, Section~\ref{sec:test-pipe} integrates these components into a unified test-time pipeline that expands memory from confident samples and applies memory-guided tree search to uncertain samples.

\vspace{-5pt}
\subsection{Step-wise memory unit: (sub-goal, sub-instruction) pairs}
\vspace{-5pt}
\label{sec:mem-unit}

A memory unit is an asymmetric pair $(g_i, u_i)$, where the \emph{sub-goal} $g_i \in \mathbb{R}^{d}$ is an embedding that summarizes the objective the sub-instruction is intended to achieve, and the \emph{sub-instruction} $u_i$ is a natural-language directive used to condition generation when the unit is applied. The memory is defined as $\mathcal{M} = \{(g_i, u_i)\}_{i=1}^{N}$, where $N$ grows over time as new reasoning experience is accumulated. We use $a_i$ to denote the action of applying unit $(g_i, u_i)$. Each unit is associated with a learnable selection head that predicts the applicability based on previous experience.

We formulate that the application of a sub-instruction depends on two factors: the objective of the next reasoning step and the conditions in the current context. The sub-goal captures the first factor and serves as a retrieval key for determining whether a sub-instruction is applicable to the current reasoning state. The learned selection heads further identify relevant conditions in the reasoning state and predict applicability scores to rerank memory units (Section~\ref{sec:mem-rs}). We organize memory as a collection of \emph{modular} units, enabling \emph{incremental} test-time expansion: new units are identified based on sub-goal similarity (Section~\ref{sec:mem-rs}), and their corresponding selection heads (Section~\ref{sec:learn-head}) can be added without retraining existing components, allowing memory to grow continuously with the test stream.

\textbf{Memory unit summarization.} Given a set of (reasoning step, step-goal) tuples collected from confident reasoning trajectories, where the step-goal describes the objective of the reasoning step and is generated by prompting the LLM, we embed the reasoning steps and step-goals using an embedding model and cluster them using k-means with the following joint distance:
\begin{equation}
    d(\mathbf{s}_{1}, \mathbf{o}_{1}; \mathbf{s}_{2}, \mathbf{o}_{2}) \;=1-\mathrm{cos}(\mathbf{s}_{1}, \mathbf{s}_{2}) \;+\; \alpha_{\text{sum}}\left[1 - \mathrm{cos}(\mathbf{o}_{1}, \mathbf{o}_{2})\right],
    \label{eq:dist-sum}
\end{equation}
where $\mathbf{s}_{1,2}$ and $\mathbf{o}_{1,2}$ denote step and step-goal embeddings, and $\alpha_{\text{sum}}$ balances the two terms. Within each cluster, the LLM summarizes a generalized sub-instruction $u_i$ from reasoning steps, while the average step-goal embedding defines the sub-goal $g_i$ of the new memory unit. Summarizing sub-instructions from similar reasoning steps promotes their generalizability, while pairing each reasoning step with its corresponding step-goal aims to ensure that the resulting sub-instruction can achieve the intended step-goal.

We also store the mean intra-cluster distance $\overline{d}_i$ to define the per-unit \emph{effective threshold} $R_{\text{eff}}^{(i)}$ (Appendix~\ref{sec:eff-range}). All embeddings of reasoning steps, step-goals, sub-goals, and context chunks are produced by the same frozen embedding model, as specified in Section~\ref{sec:reason-perf}.

\vspace{-5pt}
\subsection{Coarse-to-fine memory retrieval and selection}
\vspace{-5pt}
\label{sec:mem-rs}

Built upon sub-goal embeddings and selection heads, we propose a two-layer \emph{coarse-to-fine memory retrieval and selection} mechanism: a lightweight similarity-based stage, followed by a learned selection head that scores each candidate using features of the full reasoning state.

\textbf{Layer 1: coarse retrieval based on sub-goal similarity.} We first prompt the LLM to generate one or more candidate step-goals for the next reasoning step and compute their embeddings using the embedding model. For each candidate step-goal embedding $\hat{g}$, we compute the cosine similarity $\mathrm{cos}(\hat{g}, g_i)$ between $\hat{g}$ and each stored sub-goal $g_i$ in $\mathcal{M}$, and retrieve the most similar memory units. To avoid retrieving irrelevant units, we apply a per-unit \emph{effective threshold} $R_{\text{eff}}^{(i)} = \alpha_{\text{eff}} \cdot \overline{d}_{i}$; units with distance exceeds the threshold are excluded. If no unit matches a candidate step-goal, the LLM directly generates the next reasoning step, and the resulting (step-goal, reasoning step) pair is stored as a seed for incremental memory expansion, gradually expanding coverage over a broader range of conditions in the step-goal embedding space. Layer 1 selects memory units whose sub-instructions address step-goals most similar to that of the next step, Layer 1 thus selects by semantic relevance.

\textbf{Layer 2: learned applicability scoring.} Layer~1 does not incorporate selection knowledge learned from previous outcome feedback. Layer~2 addresses this limitation with a learned per-unit \emph{memory selection head} $\pi_{\boldsymbol{\theta}_{i}}(a_i \mid x)$ that scores the applicability of the $i$-th unit in state $x$. Specifically, we first identify contextual patterns from the input context and then use the selection heads to predict applicability scores based on these patterns. To identify contextual patterns most relevant to the applicability of $a_i$ and to obtain a fixed-length representation of a variable-length reasoning state $x$, we pre-compute a small set of \emph{context prototypes} $\{\mathbf{q}_{i1}, \ldots, \mathbf{q}_{in}\}$ by clustering embeddings of stored step-goals and contextual conditions from previous applications of the $i$-th memory unit; the full set $\{\mathbf{q}_{ij}\}$ forms a \emph{context codebook} (Appendix~\ref{sec:codebook}). Given $x$, we split it into overlapping chunks with embeddings $\mathbf{c}_1, \ldots, \mathbf{c}_T$ and apply the codebook via max-similarity pooling:
\begin{equation}
    \mathbf{m}_i(x) \;=\; [s_{i1}, s_{i2}, \ldots, s_{in}], \qquad
    s_{ij} \;=\; \max_{k \in [1, T]} \mathrm{cos}(\mathbf{q}_{ij}, \mathbf{c}_k),
    \label{eq:context-feature}
\end{equation}
where $\mathbf{m}_i(x) \in \mathbb{R}^{n}$ is the context feature vector. The max-similarity construction gives a location-invariant fingerprint of the reasoning state with respect to each prototype, regardless of how long $x$ becomes. Based on the context feature vector, each per-unit selection head scores the applicability of the $i$-th unit at state $x$: $\pi_{\boldsymbol{\theta}_{i}}(a_i \mid x) = f_{\boldsymbol{\theta}_{i}}\bigl(\mathbf{m}_i(x)\bigr) \in [0, 1]$. Each selection head is implemented as a lightweight one-vs-rest binary classifier, as detailed in Section~\ref{sec:learn-head}. Layer 2 reranks memory units using selection heads trained on previous experience, enabling selection based on learned applicability.

\textbf{Combined score.} To balance semantic relevance (Layer~1) and learned applicability (Layer~2), we combine the Layer~1 similarity score with the Layer~2 applicability score:
\begin{equation}
    \mathrm{score}(a_i, x) \;=\; (1 - \beta_{\text{c2f}}) \cdot \mathrm{cos}(\hat{g}, g_i) \;+\; \beta_{\text{c2f}} \cdot \pi_{\boldsymbol{\theta}_{i}}(a_i \mid x),
    \label{eq:combined-score}
\end{equation}
where $\beta_{\text{c2f}} \in [0, 1]$ controls the relative weight of the two layers. The memory unit that passes Layer 1 filtering and achieves the highest combined score is applied in the next reasoning step.

\textbf{Train and apply selection heads via two-layer selection.} Layer~1 selection is used on confident samples to collect sub-instruction application trajectories for training the selection heads. This enables the selection heads to learn when each sub-instruction should be applied from the LLM’s confident reasoning trajectories. For uncertain samples, Layer~1 selection first retrieves candidate memory units, after which Layer~2 selection uses the trained heads to rerank candidate sub-instructions, allowing the model to leverage knowledge learned from confident cases to guide reasoning under uncertainty.

\vspace{-5pt}
\subsection{Learning the memory selection heads}
\vspace{-5pt}
\label{sec:learn-head}

To update the selection heads for improved performance. We formulate memory selection as a sequential decision-making problem, where taking action $a_i$ (selecting the $i$-th memory unit) at reasoning state $x$ should maximize downstream final-answer correctness. We first derive the optimal selection policy under a KL regularization constraint, then approximate it with a lightweight per-unit classifier suitable for incremental test-time learning. Training data is collected from confident samples using the Layer~1 selection.

\textbf{Selection objective.} We formalize step-wise memory selection as a sequential decision-making problem and model the choice of which action $a_i$ to apply at state $x$ as a stochastic \emph{selection policy} $\pi(a_i \mid x)$ over the units in $\mathcal{M}$. Let $p(\cdot \mid \cdot)$ denote the original LLM generation distribution and $p_m(\cdot \mid \cdot)$ the memory-guided generation distribution. Our objective is to maximize the probability of producing a correct final answer under memory guidance:
\begin{equation}
    \max_{\pi(a_i \mid x)} p_m(y^*|x)=\sum\nolimits_{i} \pi(a_i \mid x) \cdot V(x, a_i),
    \quad
    V(x, a_i) \;=\; \mathbb{E}_{p(e \mid x, a_i)}\!\left[\mathbb{E}_{p_{m}(y \mid e, x)}[r]\right],
    \label{eq:final-obj}
\end{equation}
where $r$ is a binary correctness indicator, $e$ denotes the next reasoning step generated under the guidance of the sub-instruction associated with the $i$-th memory unit, and $y$ is the resulting final answer. The full derivation is provided in Appendix~\ref{sec:slt-obj-apd}.

\textbf{Closed-form optimum under a KL constraint.} Given trajectories collected by the Layer~1 data-collection distribution $\pi_{\text{sim}}(a_i \mid x)$, augmenting Eq.~\eqref{eq:final-obj} with a KL-divergence penalty $\beta_{\text{KL}}\,\mathrm{KL}\!\bigl(\pi(\cdot \mid x) \,\Vert\, \pi_{\text{sim}}(\cdot \mid x)\bigr)$ yields the following closed-form optimum \cite{peng2019advantage}:
\begin{equation}
    \pi^{*}(a_i \mid x) \;=\; \frac{1}{Z(x)} \, \pi_{\text{sim}}(a_i \mid x) \cdot \exp\!\bigl(V(x, a_i) / \beta_{\text{KL}}\bigr),
    \label{eq:opt-policy}
\end{equation}
where $\beta_{\text{KL}} > 0$ is the KL coefficient and $Z(x)$ is a normalization constant ensuring that $\pi^{*}(\cdot \mid x)$ is a valid distribution. A detailed derivation is provided in Appendix~\ref{sec:opt-teacher}. The closed-form optimum characterizes the desired sub-instruction selection behavior but does not provide a practical mechanism for generalization beyond the sampled trajectories. We therefore train lightweight selection heads to approximate $\pi^{*}(\cdot \mid x)$, enabling effective memory-selection patterns learned from rollout outcomes to be transferred to unseen reasoning states.

\textbf{From the closed-form optimum to a tractable surrogate.} Directly approximating the optimal policy in Eq.~\eqref{eq:opt-policy} is impractical because $\pi_{\text{sim}}$ lacks a tractable density and rollout-based value estimates are noisy. We therefore approximate the policy by training one-vs-rest binary classifiers on thresholded correctness labels, encouraging the selection heads to imitate successful decisions sampled from $\pi_{\text{sim}}(a_i \mid x)$. This per-unit design supports incremental memory expansion without retraining existing heads. Further discussion is provided in Appendix~\ref{sec:prac-obj}.

\textbf{Per-unit binary cross-entropy training.} For each memory unit, we train its selection head $\pi_{\boldsymbol{\theta}_{i}}(a_i \mid x)$ via binary cross-entropy on (state, label) pairs collected from MCTS rollouts (Section~\ref{sec:test-pipe}):
\begin{equation}
    \ell_{i}(x, \boldsymbol{\theta}_{i}; r_{i}) \;=\; -\left[\, r_{i} \cdot \log \pi_{\boldsymbol{\theta}_{i}}(a_i \mid x) \;+\; (1 - r_{i}) \cdot \log\!(1 - \pi_{\boldsymbol{\theta}_{i}}(a_i \mid x)) \,\right].
    \label{eq:cond-loss}
\end{equation}
In the one-vs-rest setting, $r_{i} \in \{0, 1\}$ denotes the binary label for action $a_i$. Positive examples for the $i$-th unit are states $x$ at which applying $a_i$ led to a correct final answer. Negative examples for the $i$-th unit include: (i) states where applying $a_i$ leads to an incorrect answer, and (ii) states where a different unit $i' \neq i$ led to a correct answer. This dense step-level supervision enables effective learning under limited-data in test time. A detailed comparison with prior optimization methods is presented in Appendix~\ref{sec:prac-obj}.

\begin{table}[]
    \caption{Final-answer accuracy (\%) on six reasoning benchmarks for four frozen backbones. MILES (rightmost column) matches or exceeds baseline works on all rows; per-row best in bold.}\vspace{-3pt}
    \label{tab:test-time}
    \centering
    \resizebox{\linewidth}{!}{
    \begin{tabular}{ll cccccc >{\columncolor{ourcol}}c}
    \toprule
    \rowcolor{headcol}
    Dataset & Model & ZS-CoT & SC & DC & BoT & AM & RB & \textbf{Ours} \\
    \midrule
    \multirow{2}{*}{MATH-500} & GPT-4.1-mini & \conf{87.33}{1.52} & \conf{89.20}{1.31} & \conf{84.53}{4.10} & \conf{87.47}{1.52} & \conf{83.47}{0.76} & \conf{84.67}{3.04} & \textbf{\conf{92.27}{1.03}} \\
    & GPT-4.1 & \conf{84.60}{5.60} & \conf{87.00}{2.28} & \conf{84.30}{2.73} & \conf{85.80}{2.48} & \conf{85.53}{0.29} & \conf{82.80}{0.86} & \textbf{\conf{91.67}{0.29}} \\
    \midrule
    \multirow{4}{*}{AIME 2024} & GPT-4.1-mini & \conf{48.67}{3.02} & \conf{59.67}{1.76} & \conf{62.33}{2.26} & \conf{52.67}{5.60} & \conf{42.00}{2.56} & \conf{47.33}{2.93} & \textbf{\conf{66.36}{1.38}} \\
      & GPT-4.1 & \conf{40.67}{4.02} & \conf{46.67}{2.75} & \conf{52.00}{3.41} & \conf{48.34}{1.26} & \conf{48.00}{3.41} & \conf{51.33}{4.08} & \textbf{\conf{54.66}{2.01}} \\
      & GPT-OSS-20B & \conf{86.67}{2.75} & \conf{93.33}{1.59} & \conf{39.33}{2.46} & \conf{53.34}{2.51} & \conf{72.66}{3.33} & \conf{70.67}{5.13} & \textbf{\conf{94.00}{1.88}} \\
      & Qwen3-30B-Instruct & \conf{67.33}{1.88} & \conf{80.00}{3.89} & \conf{28.00}{3.02} & \conf{66.67}{2.51} & \conf{71.33}{2.01} & \conf{70.00}{2.25} & \textbf{\conf{85.33}{1.23}} \\
    \midrule
    \multirow{4}{*}{AIME 2025} & GPT-4.1-mini & \conf{44.67}{2.56} & \conf{53.67}{2.38} & \conf{59.33}{3.14} & \conf{41.37}{3.62} & \conf{31.33}{2.56} & \conf{40.00}{5.73} & \textbf{\conf{60.33}{1.35}} \\
      & GPT-4.1 & \conf{34.00}{2.93} & \conf{38.00}{1.23} & \textbf{\conf{47.33}{2.46}} & \conf{30.67}{2.46} & \conf{36.00}{3.69} & \conf{36.67}{2.75} & \conf{42.67}{1.88} \\
      & GPT-OSS-20B & \conf{84.67}{3.41} & \conf{88.67}{1.23} & \conf{31.33}{4.93} & \conf{45.00}{8.80} & \conf{68.00}{3.02} & \conf{71.33}{2.01} & \textbf{\conf{92.66}{1.00}} \\
      & Qwen3-30B-Instruct & \conf{56.67}{1.59} & \conf{66.67}{2.25} & \conf{30.00}{2.25} & \conf{46.67}{5.03} & \conf{63.33}{3.90} & \conf{65.33}{1.23} & \textbf{\conf{72.66}{1.01}} \\
    \midrule
    \multirow{4}{*}{\parbox{2cm}{GPQA\\-Diamond}} & GPT-4.1-mini & \conf{70.71}{2.18} & \conf{72.02}{2.70} & \conf{64.82}{3.15} & \conf{66.50}{4.40} & \conf{67.34}{1.91} & \conf{66.67}{3.32} & \textbf{\conf{73.06}{1.91}} \\
      & GPT-4.1 & \conf{68.44}{3.92} & \conf{69.19}{3.32} & \conf{62.97}{7.24} & \conf{62.80}{3.63} & \conf{68.18}{3.76} & \conf{69.19}{3.76} & \textbf{\conf{70.88}{0.72}} \\
      & GPT-OSS-20B & \conf{69.36}{5.08} & \conf{71.05}{2.90} & \conf{61.95}{0.72} & \conf{37.20}{0.72} & \conf{64.98}{5.94} & \conf{62.80}{4.41} & \textbf{\conf{74.58}{2.61}} \\
      & Qwen3-30B-Instruct & \conf{70.03}{0.72} & \conf{72.89}{1.45} & \conf{42.76}{0.73} & \conf{68.01}{0.72} & \conf{70.37}{6.90} & \conf{70.03}{1.92} & \textbf{\conf{73.74}{1.25}} \\
    \midrule
    \multirow{4}{*}{\parbox{2cm}{MMLU-Pro\\Physics}} & GPT-4.1-mini & \conf{82.33}{2.59} & \conf{83.00}{1.24} & \conf{83.50}{1.24} & \conf{83.67}{1.50} & \conf{81.17}{3.79} & \conf{80.67}{1.90} & \textbf{\conf{84.83}{1.90}} \\
      & GPT-4.1 & \conf{80.50}{2.15} & \conf{82.83}{1.90} & \conf{75.83}{9.32} & \conf{83.17}{3.59} & \conf{82.83}{2.86} & \conf{83.00}{2.48} & \textbf{\conf{85.00}{1.24}} \\
      & GPT-OSS-20B & \conf{84.67}{2.59} & \conf{85.83}{1.90} & \conf{80.33}{5.74} & \conf{43.00}{2.48} & \conf{81.33}{2.59} & \conf{79.83}{5.02} & \textbf{\conf{86.83}{0.72}} \\
      & Qwen3-30B-Instruct & \conf{87.33}{0.72} & \textbf{\conf{88.50}{3.73}} & \conf{64.33}{1.90} & \conf{83.83}{3.52} & \conf{83.17}{1.90} & \conf{85.50}{3.73} & \textbf{\conf{88.50}{1.24}} \\
    \midrule
    \multirow{4}{*}{\parbox{2cm}{MMLU-Pro\\Engineering}} & GPT-4.1-mini & \conf{67.67}{4.70} & \conf{72.33}{3.13} & \conf{68.67}{5.02} & \conf{71.50}{4.30} & \conf{61.83}{5.17} & \conf{61.17}{0.72} & \textbf{\conf{75.17}{1.89}} \\
      & GPT-4.1 & \conf{73.67}{3.99} & \conf{78.50}{1.24} & \conf{67.67}{7.05} & \conf{72.17}{4.46} & \conf{68.17}{1.43} & \conf{69.67}{1.90} & \textbf{\conf{80.00}{1.24}} \\
      & GPT-OSS-20B & \conf{67.17}{9.16} & \conf{68.67}{4.90} & \conf{62.00}{7.32} & \conf{38.17}{5.42} & \conf{66.17}{3.79} & \conf{62.17}{4.36} & \textbf{\conf{71.00}{1.24}} \\
      & Qwen3-30B-Instruct & \conf{74.67}{1.90} & \conf{77.17}{1.90} & \conf{52.50}{1.24} & \conf{75.33}{0.72} & \conf{79.00}{5.69} & \conf{75.50}{2.15} & \textbf{\conf{80.83}{3.73}} \\
    \bottomrule
    \end{tabular}} \vspace{-7pt}
\end{table}

\vspace{-5pt}
\subsection{Incremental test-time memory construction and inference}
\vspace{-5pt}
\label{sec:test-pipe}

In this section, we integrate the previous components into a unified pipeline, including: (1) incrementally constructing memory and learning sub-instruction selection from confident samples, and (2) applying memory-guided reasoning to uncertain samples.

\textbf{Confidence-based routing of test samples.} In the test-time setting, samples arrive sequentially without ground-truth labels. For each test sample $q$, we generate $K$ independent LLM responses and check whether they agree on a final answer. Samples whose responses converge to a consistent answer are treated as \emph{confident samples}: the majority-vote answer serves as a pseudo-label for evaluating MCTS rollouts, and the resulting trajectories are used to expand memory and train the selection heads. Samples with inconsistent responses are instead routed to memory-guided tree search using the learned heads. Intuitively, confident samples provide a reliable supervisory signal, whereas the LLM should leverage prior experience to guide reasoning when it is uncertain about a sample.

\textbf{Data collection and memory expansion via MCTS on confident samples.} For each confident sample we run MCTS over the reasoning tree to collect training data for the selection heads. At each tree node (a reasoning state $x$), we use the Layer~1 sub-goal similarity to sample candidate memory units to apply, run the LLM to expand the chosen step, and propagate correctness signals from rollouts evaluated against the pseudo-label. The detailed MCTS procedure is presented in Appendix~\ref{sec:mcts-detail}.

The collected (step-goal, reasoning step) pairs (Section~\ref{sec:mem-rs}) are summarized into new memory units as described in Section~\ref{sec:mem-unit}. Data collection uses only Layer~1 similarity-based sampling via $\pi_{\text{sim}}$ to avoid bias toward the current selection heads and preserve trajectory diversity. The resulting tuples are stored in per-unit datasets $\mathcal{D}_{i} = \{(x_{j}, r_{i,j})\}$ for training the corresponding selection heads $\pi_{\boldsymbol{\theta}_{i}}$ with the BCE loss in Eq.~\eqref{eq:cond-loss}. The detailed training procedure is described in Appendix~\ref{sec:detail-train}.

\textbf{Memory-guided tree search for uncertain samples.} For samples whose initial $K$ responses are inconsistent, we apply a multi-stage memory-guided tree search similar to \cite{qiu2024treebon}. At each stage, a sub-context is sampled from existing reasoning chains, candidate step-goals are generated, and memory units are retrieved and ranked using the combined score in Eq.~\eqref{eq:combined-score}. The top-ranked unit is then applied to expand the reasoning trajectory, and the resulting branch is added to the response pool. As search progresses, longer sub-contexts are used to guide later reasoning steps. The final answer is determined by majority voting over all generated responses. Details of the memory-guided tree search procedure are provided in Appendix~\ref{sec:mgts-detail}. 

\vspace{-5pt}
\section{Experiments}
\vspace{-5pt}

\subsection{Main results on LLM reasoning: accuracy, efficiency, and transferability}
\vspace{-5pt}
\label{sec:reason-perf}

\textbf{Datasets and models evaluation.} We evaluate on six reasoning benchmarks spanning competition mathematics and graduate-level academic exams: MATH-500 (500 samples)~\cite{hendrycks2021measuring}, AIME 2024 (30 samples), AIME 2025 (30 samples), GPQA-Diamond (198 samples)~\cite{rein2024gpqa}, and the Physics and Engineering subsets of MMLU-Pro~\cite{wang2024mmlu}. Following~\cite{suzgun2025dynamic}, we randomly sample 200 instances from each MMLU-Pro subset and evaluate the remaining datasets in full. We evaluate MILES on four backbone LLMs spanning proprietary and open-weight families: GPT-4.1, GPT-4.1-mini, Qwen3-30B-Instruct~\cite{yang2025qwen3}, and GPT-OSS-20B~\cite{agarwal2025gpt}. For fair comparison, the same backbone is used for both reasoning and memory construction across all methods within each row. All embeddings in MILES are computed using the embedding model \texttt{text-embedding-3-small}. Final-answer accuracy is used as the evaluation metric. 

\begin{wrapfigure}{R}{0.6\textwidth}
    \centering
    \vspace{-13pt}
    \includegraphics[width=0.95\linewidth]{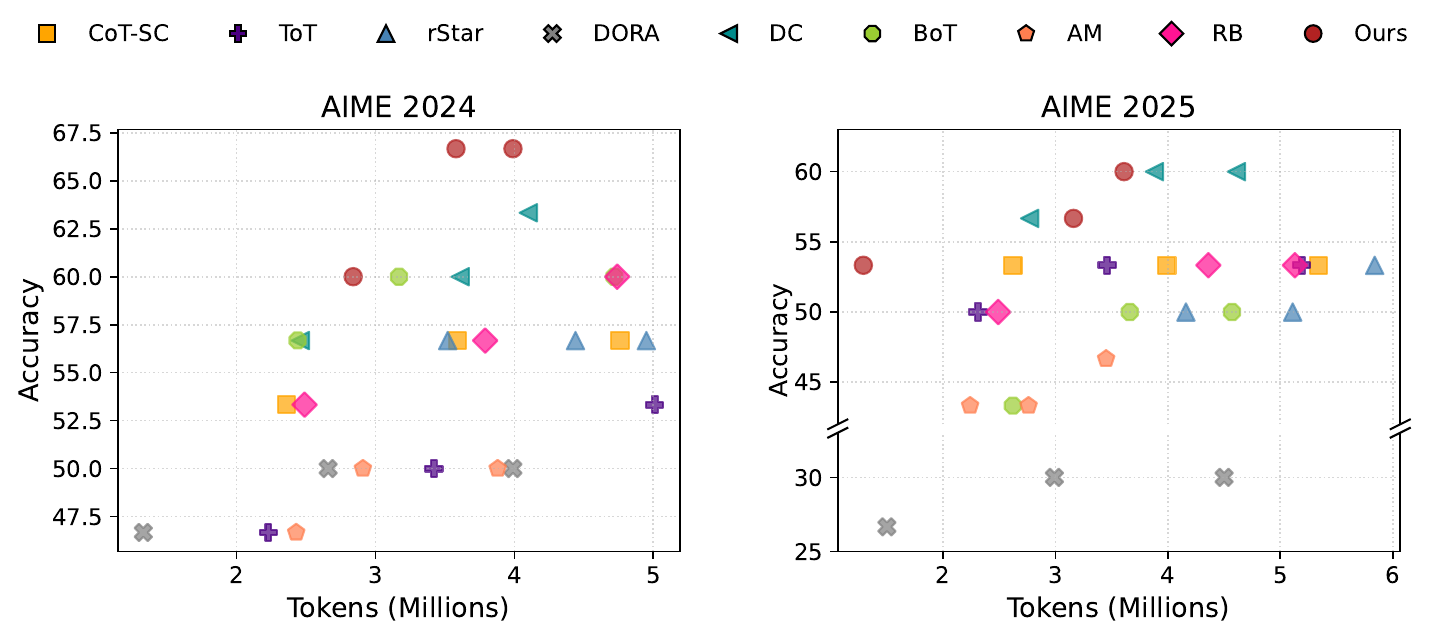}\vspace{-7pt}
    \caption{Accuracy vs.\ response-token cost on AIME 2024 and AIME 2025 with GPT-4.1-mini, comparing MILES with test-time scaling and memory-based baselines. MILES reaches the highest accuracy on AIME 2024 and ties the strongest baseline, DC, on AIME 2025.} 
    \vspace{-10pt}
    \label{fig:acc-vs-token}
\end{wrapfigure}
\textbf{Evaluation:} In the test-time setting, samples arrive sequentially, and memory is constructed only from solutions to previously processed samples. Predictions for previous samples cannot be revised and all methods are evaluated on the same sequence of questions. We evaluate final-answer accuracy across multiple runs, with numbers in parentheses indicating 95\% confidence intervals.

\textbf{Baselines.} We compare MILES against test-time memory methods, including Buffer-of-Thoughts (BoT)~\cite{yang2024buffer} and Dynamic CheatSheet (DC)~\cite{suzgun2025dynamic}, ArcMemo (AM) \cite{ho2025arcmemo} and ReasoningBank (RB) \cite{ouyang2025reasoningbank}, as well as two standard baselines: zero-shot Chain-of-Thought (ZS-CoT)~\cite{kojima2022large} and Self-Consistency (SC)~\cite{wang2022self}. Additional comparisons with memoryless test-time scaling methods are provided in Appendix~\ref{sec:additional-exp}.

\textbf{MILES configuration.} Across all experiments, MILES uses the same component design described in Section~\ref{sec:method}, with per-unit selection heads implemented as an ensemble of 10 independently trained 3-layer MLPs for robust inference. Additional implementation details are provided in Appendix~\ref{sec:exp-detail}. 


Table~\ref{tab:test-time} shows that MILES matches or outperforms baseline works across all six benchmarks and four backbones, with consistent gains on both math (MATH-500, AIME 2024--2025) and academic (GPQA-Diamond, MMLU-Pro Physics/Engineering) datasets. The improvements are especially pronounced on smaller open-weight backbones (GPT-OSS-20B and Qwen3-30B-Instruct), where the sample-level memory baselines BoT~\cite{yang2024buffer} and DC~\cite{suzgun2025dynamic} fail to outperform zero-shot CoT due to overly complex prompt. In near-saturated settings (e.g., GPT-OSS-20B on AIME 2024), MILES matches self-consistency without degradation, demonstrating its effectiveness and robustness. We note that the strong performance of DC on the AIME datasets mainly comes from its use of coding; additional details are provided in Appendix~\ref{sec:additional-exp}.

\begin{wraptable}{R}{0.5\textwidth}
\centering
\setlength{\tabcolsep}{4pt}
\vspace{-12pt}
\caption{Cross-model transfer on AIME 2024 and AIME 2025 with GPT-4.1-mini as the reasoning backbone. Memory built from rollouts of an auxiliary model (Qwen3-30B-Instruct, Qwen3-4B-Instruct, or GPT-4.1-mini itself) improves over the no-auxiliary baseline (top row in each block).}\vspace{-5pt}
\label{tab:tranfer}
    \begin{tabular}{ll >{\columncolor{ourcol}}c}
    \toprule
    \rowcolor{headcol}
     Dataset & Auxiliary model & Accuracy \\
     \midrule
    \multirow{4}{*}{AIME 2024} & no-auxiliary model & \conf{59.33}{1.00} \\
     & Qwen3-30B-Instruct & \textbf{\conf{68.00}{2.01}} \\
     & Qwen3-4B-Instruct & \conf{64.67}{1.23} \\
     & GPT-4.1-mini & \conf{66.36}{1.38} \\
     \midrule
     \multirow{4}{*}{AIME 2025} & no-auxiliary model & \conf{53.33}{1.59} \\
     & Qwen3-30B-Instruct & \conf{60.00}{1.12} \\
     & Qwen3-4B-Instruct & \conf{57.33}{1.88} \\
     & GPT-4.1-mini & \textbf{\conf{60.33}{1.35}} \\
     \bottomrule
    \end{tabular}
\vspace{-10pt}
\end{wraptable}
\textbf{Performance under different computation budget.} We probe how accuracy scales with test-time compute by plotting accuracy against response-token cost on AIME 2024 and AIME 2025 with GPT-4.1-mini, comparing MILES to representative memoryless test-time scaling methods: Self-Consistency~\cite{wang2022self}, Tree-of-Thoughts~\cite{yao2023tree}, rStar~\cite{qi2024mutual}, and DORA~\cite{wang2025every}. As shown in Figure~\ref{fig:acc-vs-token}, MILES benefits more from increased computation budgets by constructing additional memory units (Section~\ref{sec:rollouts}), dominating the accuracy/token frontier and continuing to improve as the budget grows, whereas the baselines quickly plateau. This highlights the effectiveness of reusing cross-problem memory, which captures knowledge that per-problem scaling alone cannot.

\textbf{Cross-model transfer.} \label{sec:transfer} Because MILES operates on textual memory representations, memory construction need not use the same model as the reasoning backbone. We evaluate cross-model transfer on AIME 2024 and AIME 2025 using GPT-4.1-mini as the reasoning backbone: the backbone generates the first 10 self-consistency rollouts for routing, while an \emph{auxiliary} model generates the next 40 rollouts for memory construction. Table~\ref{tab:tranfer} shows that auxiliary trajectories from Qwen3-30B-Instruct, Qwen3-4B-Instruct, and GPT-4.1-mini itself all improve over the no-auxiliary baseline, demonstrating that memory constructed by open-weight auxiliary models transfers effectively to a stronger frozen backbone. 

\begin{wraptable}{R}{0.5\textwidth}
\centering
\setlength{\tabcolsep}{2.8pt}
\vspace{-12pt}
\caption{Memory pre-built by an auxiliary model on a separate training set with ground-truth labels (AIME 2020--2024 for AIME 2025; non-Diamond GPQA for GPQA-Diamond), evaluated on the most uncertain test samples with GPT-4.1-mini as the reasoning backbone. Pre-built memory consistently improves over the no-memory baseline (top row in each block).}\vspace{-5pt}
\label{tab:train-tranfer}
    \begin{tabular}{ll >{\columncolor{ourcol}}c}
    \toprule
    \rowcolor{headcol}
     Dataset & Auxiliary model & Accuracy \\
     \midrule
    \multirow{3}{*}{AIME 2025} & no memory & \conf{18.46}{2.84} \\
     & Qwen3-30B-Instruct & \conf{28.46}{2.66} \\
     & GPT-4.1-mini & \textbf{\conf{36.92}{2.32}} \\
     \midrule
     \multirow{3}{*}{\parbox{2cm}{GPQA-\\Diamond}} & no memory & \conf{45.97}{2.39} \\
     & Qwen3-30B-Instruct & \textbf{\conf{56.77}{1.19}} \\
     & GPT-4.1-mini & \conf{54.19}{0.97} \\
     \bottomrule
    \end{tabular}
\end{wraptable}
\textbf{Offline memory construction with training set.} \label{sec:train-set} When a separate labeled set is available, memory can be built ahead of time and memory-guided tree search applied only to the most uncertain test samples (Section \ref{sec:test-pipe}). We evaluate our method on AIME 2025 using memory constructed from AIME 2020--2024, and on GPQA-Diamond~\cite{rein2024gpqa} using memory constructed from the non-Diamond GPQA split, reporting accuracy on the 13 and 31 most uncertain test samples, respectively. Table~\ref{tab:train-tranfer} shows that memory pre-built by either Qwen3-30B-Instruct or GPT-4.1-mini itself substantially improves over the no-memory baseline. These results demonstrate that MILES learns reusable problem-solving knowledge that transfers across different models and samples within the same domain.

\vspace{-5pt}
\subsection{Ablation study}
\vspace{-5pt}
\label{sec:ablation}

\textbf{Effectiveness of each component in two-layer selection.} To evaluate the two-layer selection mechanism, we conduct experiments on AIME 2024 and AIME 2025 with GPT-4.1-mini. When both layers are disabled, no memory is constructed and no selection heads are trained. We further compare Layer~2 against two alternative rerankers: (i)~\emph{state-similarity reranking}, which scores candidates by the similarity between the current reasoning-state embedding and the average embeddings of previously successful states; and (ii)~\emph{LLM-prompt reranking}, which prompts LLM with the current reasoning state and the Layer-1 candidate sub-instructions and asks it to identify the most applicable candidate.

\begin{wraptable}{r}{0.55\textwidth}
\centering
\setlength{\tabcolsep}{4pt}
    \caption{Ablation study of memory selection components. Incorporating selection heads leads to consistent performance improvements.}\vspace{-5pt}
    \label{tab:ablation}
    \resizebox{\linewidth}{!}{
    \begin{tabular}{lccc >{\columncolor{ourcol}}c}
    \toprule
    \rowcolor{headcol}
         Dataset & Exp. ID & Layer 1 & Layer 2 & Accuracy \\
    \midrule
         \multirow{6}{*}{\parbox{1.2cm}{AIME\\ 2024}} & 1 & \xmark & \xmark & \conf{59.33}{1.00} \\
         & 2 & \gmark & \xmark & \conf{62.66}{1.01} \\
         & 3 & \xmark & \gmark & \conf{64.00}{1.88} \\
         & 4 & \gmark & \gmark & \textbf{\conf{66.36}{1.38}} \\
         \cmidrule{2-5}
         & 5 & \gmark & State sim. & \conf{64.00}{1.01} \\
         & 6 & \gmark & LLM rerank   & \conf{64.67}{2.01} \\
         \midrule
         \multirow{6}{*}{\parbox{1.2cm}{AIME\\ 2025}} & 1 & \xmark & \xmark & \conf{53.33}{1.59} \\
         & 2 & \gmark & \xmark & \conf{55.33}{2.01} \\
         & 3 & \xmark & \gmark & \conf{58.00}{1.23} \\
         & 4 & \gmark & \gmark & \textbf{\conf{60.33}{1.35}} \\
         \cmidrule{2-5}
         & 5 & \gmark & State sim. & \conf{56.00}{1.88} \\
         & 6 & \gmark & LLM rerank   & \conf{56.67}{1.60} \\
    \bottomrule
    \end{tabular}} \vspace{-15pt}
\end{wraptable}
Table~\ref{tab:ablation} shows that combining both layers (Exp~4) yields the best accuracy on both datasets, while disabling either layer (Exp~2--3) reduces performance. Disabling both layers causes accuracy to fall close to the self-consistency baseline. Both alternative rerankers (Exp~5 and Exp~6) outperform the no-reranker baseline (Exp~2). LLM-prompt reranking (Exp~6) slightly outperforms state-similarity reranking (Exp~5), but both perform worse than the learned classifier (Exp~4). This suggests that the learned classifier captures additional information beyond what LLM prompting or state similarity can provide. These results further demonstrate the effectiveness of our two-layer selection mechanism.

\begin{wrapfigure}{R}{0.6\textwidth}
    \centering
    \vspace{-10pt}
    \includegraphics[width=0.95\linewidth]{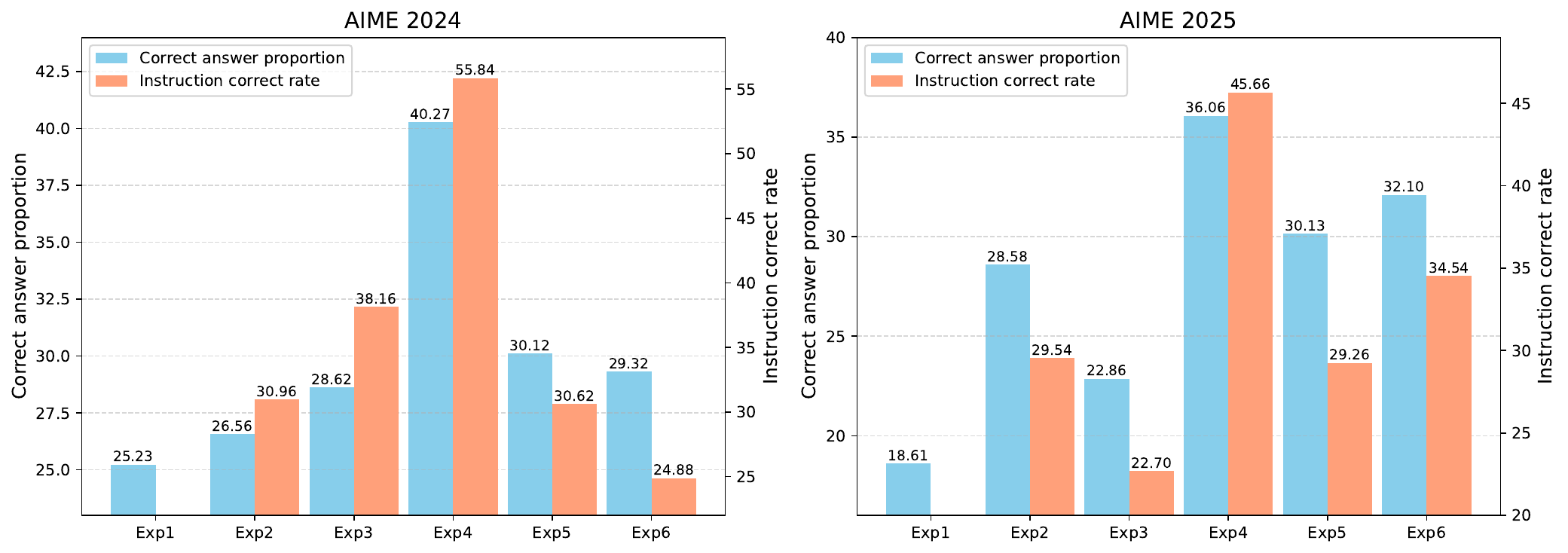}\vspace{-2pt}
    \caption{Correct-answer proportion and sub-instruction accuracy on uncertain samples for each experiment in Table~\ref{tab:ablation}. Our method (Exp~4) achieves the best performance on both.}  \vspace{-5pt}
    \label{fig:path-eval}
\end{wrapfigure}

We further inspect reasoning trajectories using two metrics: \emph{correct-answer proportion} (fraction of trajectories that produce the correct final answer) and \emph{sub-instruction accuracy} (fraction of sub-instruction applications that lead to a correct final answer), on samples near the correct/incorrect decision boundary where small effects matter most. Figure~\ref{fig:path-eval} shows that our full method (Exp~4) performs best on both metrics. The gaps between Exp~4 and Exp~1--3 validate the asymmetric memory and two-layer selection design, as well as between Exp~4 and Exp~5--6 validate the learned heads over alternative rerankers.

\textbf{Why both selection layers matter.} The results in Table~\ref{tab:ablation} demonstrate the importance of combining Layer 1 and Layer 2 for effective memory reuse. With only semantic similarity-based selection (Exp~2), a sub-instruction may be relevant to the current reasoning state but may not lead to a correct final answer when applied. Conversely, relying solely on learned selection heads (Exp~3) cannot effectively filter out semantically unrelated candidates. Combining both layers (Exp~4) achieves the best performance by integrating semantic relevance with feedback from previous experience. The improvements in both sub-instruction accuracy and correct-answer proportion (Figure~\ref{fig:path-eval}) further support the importance of learning when to apply retrieved knowledge, rather than relying solely on semantic relevance.

\begin{wraptable}{R}{0.5\textwidth}
\centering
\setlength{\tabcolsep}{4pt}
\vspace{-12pt}
\caption{Effect of the per-confident-sample MCTS rollout budget on memory size and accuracy, evaluated on AIME 2024 and AIME 2025 with GPT-4.1-mini. More rollouts produce a richer memory and higher accuracy.}\vspace{-5pt}
\label{tab:mcts-rollouts}
    \begin{tabular}{lccc}
    \toprule
    Dataset & Rollouts & Mem. size & Accuracy \\
    \midrule
    \multirow{3}{*}{AIME 2024} & 20 & 9 & \conf{60.67}{1.00} \\
    & 30 & 15 & \conf{66.00}{1.01} \\
    & 40 & 20 & \conf{66.36}{1.38} \\
    \midrule
    \multirow{3}{*}{AIME 2025} & 20 & 10 & \conf{51.33}{2.01} \\
    & 30 & 19 & \conf{58.00}{1.23} \\
    & 40 & 25 & \conf{60.33}{1.35} \\
    \bottomrule
    \end{tabular}
\vspace{-18pt}
\end{wraptable}
\textbf{Effect of MCTS rollouts.} \label{sec:rollouts} MILES summarizes sub-instructions in memory and collects classifier-training data via MCTS on confident samples; we therefore study how the rollout budget affects both the resulting memory size and the final accuracy. We sweep rollouts $\in \{20, 30, 40\}$ on AIME 2024 and AIME 2025 with GPT-4.1-mini and report results in Table~\ref{tab:mcts-rollouts}.

More rollouts yield both a larger memory with more sub-instruction units and higher accuracy. A larger number of sub-instructions corresponds to better coverage of the step-goal embedding space (Section~\ref{sec:mem-rs}). These results suggest that improved coverage leads to higher performance, and further demonstrate the effectiveness of our memory selection strategy.

\begin{wraptable}{R}{0.55\textwidth}
\centering
\vspace{-13pt}
\setlength{\tabcolsep}{4pt}
\caption{Performance of our method on the AIME 2024 and AIME 2025 datasets under different sample orders and MCTS rollout budgets using GPT-4.1-mini.}\vspace{-5pt}
\label{tab:sample-order}
\resizebox{\linewidth}{!}{
    \begin{tabular}{lccc}
    \toprule
    \multirow{2}{*}{Dataset} & \multirow{2}{*}{Sample order} & \multicolumn{2}{c}{Accuracy} \\
    \cmidrule{3-4}
    & & 20 rollouts & 40 rollouts \\
    \midrule
    \multirow{3}{*}{AIME 2024} & default & \conf{60.67}{1.00} & \conf{66.36}{1.38} \\
     & easy-to-hard & \conf{66.00}{1.01} & \conf{66.34}{0.76} \\
     & hard-to-easy & \conf{64.00}{2.46} & \conf{67.00}{0.75} \\
    \midrule
    \multirow{3}{*}{AIME 2025} & default & \conf{51.33}{2.01} & \conf{60.33}{1.35} \\
     & easy-to-hard & \conf{56.67}{1.59} & \conf{59.67}{0.75} \\
     & hard-to-easy & \conf{55.33}{2.01} & \conf{63.00}{1.35} \\
    \bottomrule
    \end{tabular}}
\vspace{-12pt}
\end{wraptable}
\textbf{Impact of sample order.} Because MILES updates memory online from confident samples (Section~\ref{sec:test-pipe}), sample order affects how knowledge is accumulated and reused. To evaluate the robustness of MILES under different sample orders, we conduct experiments on AIME 2024 and AIME 2025 with GPT-4.1-mini under three sample orders: the default question-index order, easiest-to-hardest, and hardest-to-easiest, using two MCTS rollout budgets (20 and 40 per confident sample). Sample difficulty is defined by the proportion of correct answers across self-consistency rollouts. Results are in Table~\ref{tab:sample-order}.

Under the limited rollout budget (20 per sample), the easy-to-hard order achieves the highest accuracy on both datasets, suggesting that solving easier samples first allows MILES to accumulate useful memory before encountering harder ones. With 40 rollouts, the performance gap largely disappears, indicating that the method becomes effectively order-insensitive once sufficient confident-sample data has been collected.

\textbf{Experience accumulation and robustness.} Tables~\ref{tab:mcts-rollouts} and~\ref{tab:sample-order} reveal two related effects of increasing the rollout budget. First, additional rollouts allow MILES to construct more memory units, improving the coverage of reusable reasoning experience. Second, increasing the rollout budget reduces sensitivity to the order in which samples are processed. Under a limited budget, processing easier samples first can provide useful experience for subsequent problems. With sufficient rollouts, MILES becomes less dependent on this ordering. These observations suggest that richer experience accumulation can improve both reasoning performance and robustness in the online setting.

\textbf{Additional design ablations and analyses} are provided in Appendix~\ref{sec:additional-exp}.

\vspace{-5pt}
\section{Conclusion}
\vspace{-5pt}
\label{sec:conclusion}

In this work, we propose MILES, a framework for test-time self-improvement that incrementally accumulates and composes reusable step-wise reasoning experience. By learning memory-selection patterns from confident trajectories and transferring them to uncertain reasoning states, MILES enables correctness-optimized memory composition without parameter updates, external supervision, or fixed action spaces. Extensive experiments across multiple reasoning benchmarks demonstrate its effectiveness, robustness, and transferability. These results suggest that learning reusable memory-selection knowledge is a promising direction for building continually improving reasoning systems.

\textbf{Limitations and future work.} MILES assumes a stream of related problems with reusable reasoning structure, as memory is accumulated from previously processed samples. Consequently, its benefits depend on the available experience and may vary with sample order, as examined in Section~\ref{sec:ablation}. Moreover, MILES relies on the backbone model's ability to generate reliable pseudo-labels through majority agreement. When the backbone performs poorly, agreement may not indicate correctness, potentially introducing incorrect supervision. Although our experiments demonstrate the reliability of pseudo-labels on the evaluated benchmarks (Appendix~\ref{sec:pseudo-label-quality}), their reliability on more challenging tasks may depend on the capability of the backbone and remains to be investigated.

MILES also relies on the instruction-following capability of the underlying LLM and requires rollout-based data collection, which introduces additional computational overhead. This cost can be reduced by using auxiliary models for trajectory generation (Section~\ref{sec:transfer}). Future work includes combining MILES with self-refinement methods \cite{ouyang2025reasoningbank,shinn2023reflexion} to improve instruction quality and with hierarchical memory \cite{yang2025reasonflux,cai2025flex} to improve selection efficiency.

\newpage

\section*{Acknowledgements}

This research was partially supported by the Australian Research Council through a Discovery Project grant (DP260103379) and a DECRA Fellowship (DE230101591) awarded to D. Gong.

\bibliographystyle{abbrv}
\bibliography{reference}

\newpage
\appendix

\section{Objective formulation and optimization}
\label{sec:all-derivation}

\subsection{Memory retrieval objective}
\label{sec:slt-obj-apd}

Inspired by the formulations of prior works \cite{lewis2020retrieval,xie2021explanation}, we formalize sub-instruction selection as a sequential decision-making problem, where the objective is to maximize the probability of generating the correct solution by optimizing the instruction selection function. At each step, with $a_i$ denoting the action of applying the $i$-th memory unit, we model its effect on generating the next reasoning step $e$ as
\begin{equation}
    p_m(e|x)=\sum_{i} p(e|x,a_i)\pi(a_i\mid x)=\mathbb{E}_{\pi(a_i\mid x)}\left[p(e\mid x,a_i)\right], \label{eq:step-gen-apd}
\end{equation}
where $\pi(a_i|x)$ denotes the memory selection function conditioned on the reasoning state $x$, $p(\cdot \mid \cdot)$ is the generation probability of LLM, and $p_m(\cdot \mid \cdot)$ denotes the generation probability conditioned on sub-instructions. The reasoning state $x$ consists of the question $q$ and the reasoning trajectory generated so far.\

Under the causal decoding mechanism, subsequent reasoning steps are conditioned on the current reasoning state and the newly generated step. Thus, the probability of generating the final solution $y$ can be written as $p_m(y|x,e)$. By the chain rule of probability,
\begin{equation}
    p_m(y,e|x)=p_m(y|e,x)\cdot p_m(e|x), \label{eq:prob-chain}
\end{equation}
which allows us to express the probability of generating the final solution as
\begin{equation}
    \begin{split}
        p_m(y\mid x)=&\int p_m(y,e\mid x)de \\
        =&\int p_m(y\mid e,x)\cdot p_m(e\mid x)de \\
        =&\int p_m(y\mid e,x)\left(\sum_{i} p(e\mid x,a_i)\pi(a_i\mid x)\right)de \\
        =&\sum_i \left(\pi(a_i\mid x)\int p_m(y\mid e,x)\cdot p(e\mid x,a_i)de\right) \\
        =&\sum_i \pi(a_i\mid x)\cdot \mathbb{E}_{p(e\mid x,a_i)}[p_m(y\mid e,x)].
    \end{split} \label{eq:gen-form-apd}
\end{equation}
Our objective is to maximize the probability of generating the correct final solution $y^*$ by optimizing $\pi(a_i\mid x)$, while keeping the LLM parameters fixed. Therefore, our optimization objective is
\begin{equation}
    \max_{\pi(a_i\mid x)}\ p_m(y^*\mid x)=\sum_i \pi(a_i\mid x)\cdot \mathbb{E}_{p(e\mid x,a_i)}[p_m(y^*\mid e,x)]. \label{eq:gen-obj-apd}
\end{equation}

The effect of selecting an instruction at the current step propagates through subsequent reasoning steps and influences the final answer. However, $p_m(y^*\mid e,x)$ is intractable in practice. Instead of explicitly modeling future memory selections, we approximate their impact using the expected correctness of the final answer, following the standard reinforcement learning practice of using rollout returns to evaluate intermediate decisions. Therefore, we propose to approximate $p_m(y^*\mid e,x)$ using a binary correctness indicator $r\in \{0,1\}$, defined as
\begin{equation}
    r=\begin{cases}
    1, & y=y^*,\\
    0, & y\neq y^*.
    \end{cases}
\end{equation}
Because $r$ follows a Bernoulli distribution with success probability $p_m(y^*\mid e,x)$, we approximate $p_m(y^*\mid e,x)$ by the expectation over $r$,
\begin{equation}
    p_m(y^*\mid e,x)=\mathbb{E}_{p_m(y\mid e,x)}[r].
\end{equation}
This approximation reduces the sequential decision problem to evaluating the expected return of a one-step decision, while implicitly accounting for future reasoning through rollout-based reward signals. Substituting this into the objective in Eq.~\eqref{eq:gen-obj-apd} yields
\begin{equation}
    \max_{\pi(a_i\mid x)} p_m(y^*\mid x)=\sum_{i} \pi(a_i\mid x)\cdot V(x,a_i),\quad V(x,a_i)=\mathbb{E}_{p(e\mid x,a_i)}\left[\mathbb{E}_{p_m(y\mid e,x)}[r]\right], \label{eq:final-obj-apd}
\end{equation}
where $V(x,a_i)$ represents the expected correctness of the final answer when selecting action $a_i$ under reasoning state $x$.

\subsection{Optimal teacher policy}
\label{sec:opt-teacher}

During data collection, we select sub-instructions based on the similarity between step-goal embeddings and sub-goal. Let this sampling policy be denoted as $\pi_{\text{sim}}$. Following \cite{peng2019advantage}, we optimize a target policy $\pi(a_i\mid x)$ under a KL-regularized objective:
\begin{equation}
    \begin{split}
        \arg \max_{\pi(a_i\mid x)}& \ \int d_{\pi_{\text{sim}}}(x)\left(\sum_{i} \pi(a_i\mid x)\cdot V(x,a_i)\right)dx \\
        \mathrm{s.t.}& \ \mathrm{D_{\text{KL}}}(\pi(\cdot \mid x)||\pi_{\text{sim}}(\cdot \mid x))\le \epsilon,\quad \forall \ x \\
        &\ \sum_{i} \pi(a_i\mid x)=1,\quad \forall \ x.
    \end{split} \label{eq:opt-obj-apd}
\end{equation}
where $d_{\pi_{\text{sim}}}(x)$ denotes the state distribution induced by $\pi_{\text{sim}}$, and $\pi(a_i\mid x)$ is the policy to be optimized.\

Following the derivation in \cite{peng2019advantage}, we form the corresponding Lagrangian:
\begin{equation}
    \begin{split}
        \mathcal{L}(\pi,\beta_{\text{KL}},\alpha_{\text{prob}})=&\int d_{\pi_{\text{sim}}}(x)\left(\sum_i \pi(a_i\mid x)\cdot V(x,a_i)\right)dx \\
        &+\beta_{\text{KL}}\left(\epsilon-\int d_{\pi_{\text{sim}}}(x)\cdot \mathrm{D_{\text{KL}}}(\pi(\cdot \mid x)||\pi_{\text{sim}}(\cdot \mid x))dx\right) \\
        &+\int \alpha_{x}\left(1-\sum_{i}\pi(a_i\mid x)\right)dx,
    \end{split}\label{eq:lagrangian}
\end{equation}
where $\beta_{\text{KL}}$ and $\alpha_{\text{prob}}=\{\alpha_{x}|\forall x\}$ are Lagrange multipliers for the KL and normalization constraints, respectively. Setting $\partial \mathcal{L} / \partial \pi(a_i\mid x) = 0$ yields the optimal policy:
\begin{equation}
    \pi^*(a_i\mid x)=\frac{1}{Z(x)}\pi_{\text{sim}}(a_i\mid x)\exp\left(V'(x,a_i)\right),\quad V'(x,a_i)=\frac{1}{\beta_{\text{KL}}}V(x,a_i),\label{eq:opt-policy-apd}
\end{equation}
where $Z(x)$ is the normalization constant:
\begin{equation}
    Z(x)=\exp\left(\frac{\alpha_{x}}{d_{\pi_{\text{sim}}}(x)\beta_{\text{KL}}}+1\right) =\sum_i \left(\pi_{\text{sim}}(a_i\mid x)\exp\left(\frac{1}{\beta_{\text{KL}}}V(x,a_i)\right)\right),
\end{equation}
ensuring that $\pi^*(a_i\mid x)$ is a valid probability distribution. Notably, DPO \cite{rafailov2023direct} derives a similar form of the optimal policy via a different formulation. The optimal form $\pi^{*}(a_i \mid x)$ is invariant to subtracting a state-dependent baseline from $V(x, a_i)$ in the exponent, since any such baseline is absorbed into $Z(x)$; we therefore use the bare value $V(x, a_i)$ rather than an advantage. 

\subsection{Practical optimization objective}
\label{sec:prac-obj}

\textbf{Training objective.} Our goal is to train the selection heads to approximate the optimal policy in Eq.~\eqref{eq:opt-policy-apd}. However, direct optimization using the weighted maximum-likelihood approach of \cite{peng2019advantage} is impractical because the sampling policy $\pi_{\text{sim}}(a_i\mid x)$ is intractable and rollout-based value estimates become noisy when amplified by small $\beta_{\text{KL}}$. We therefore approximate the optimal policy using binary correctness labels, where $r_i=1$ if selecting action $a_i$ leads to the correct final answer, and $r_i=0$ otherwise. 

In Eq.~\eqref{eq:opt-policy-apd}, the optimal policy $\pi^*(a_i\mid x)$ reweights the sampling policy $\pi_{\text{sim}}(a_i\mid x)$ according to the value $V(x,a_i)$. Approximating $\pi^*(a_i\mid x)$ therefore encourages the selection heads to assign higher probabilities to action $a_i$ selected by $\pi_{\text{sim}}(a_i\mid x)$ that lead to high-value outcomes and lower probabilities to those that lead to poor outcomes. Our binary surrogate follows the same intuition by training the selection heads to distinguish successful trajectories sampled from $\pi_{\text{sim}}(a_i\mid x)$ from unsuccessful ones.

\textbf{Per-unit selection head training.} To support incrementally expanding memory, we train a separate selection head for each sub-instruction rather than a single joint classifier. This design allows newly added instructions to be trained independently without retraining existing classifiers. Specifically, we decompose selection into one-vs-rest binary classification problems. For the $i$-th memory unit, positive samples are states where applying $a_i$ leads to a correct final answer, while negative samples include states where applying $a_i$ leads to an incorrect answer or where another instruction succeeds instead. We train each classifier using binary cross-entropy:
\begin{equation}
    \ell(x,\boldsymbol{\theta}_i;r_i)=-\left[r_i\cdot \log\left(\pi_{\boldsymbol{\theta}_i}(a_i\mid x)\right)+(1-r_i)\cdot \log\left(1-\pi_{\boldsymbol{\theta}_i}(a_i\mid x)\right)\right], \label{eq:bce-loss}
\end{equation}
where $r_i\in\{0,1\}$ denotes the binary label for action $a_i$. This objective encourages high scores for instructions that lead to correct solutions while suppressing unsuitable ones.\

\textbf{Collecting diverse training data.} In our data collection procedure, we select sub-instructions based on the similarity between step-goal embeddings and sub-goal, without relying on predictions from the selection heads. This design aims to mitigate selection bias and promote data diversity. In contrast, using the selection head during data collection would bias the sampled applications toward its current preferences, resulting in less informative and less diverse training data. By decoupling data collection from the selection head, we obtain a more diverse set of application trajectories, which enables the selection head to more effectively learn when to apply each sub-instruction.\

\textbf{Connection to and departure from prior derivations.} The closed-form Eq.~\eqref{eq:opt-policy-apd} we use is the same form as the optimum derived in \cite{peng2019advantage} for advantage-weighted regression and, in a different domain, in \cite{rafailov2023direct} for direct preference optimization. We share with these works only the \emph{derivation of the optimum}. 

In AWR \cite{peng2019advantage}, the policy is iteratively improved by applying the current policy and updating it toward a new optimal policy. However, in our setting, iteratively refining the policy for a single question may introduce bias and degrade performance on other questions, while updating the policy across all existing questions is computationally expensive. In practice, we replace AWR's soft advantage-weighted maximum-likelihood surrogate with a binary cross-entropy surrogate over per-unit classifiers, and we do not perform iterative policy improvement: training data is collected once via MCTS rollouts on confident samples (Section~\ref{sec:test-pipe}), and the per-unit heads are trained on the collected data. We therefore avoid iterative policy updates.

ExIt \cite{anthony2017thinking} also employs iterative updates: in each iteration, it collects trajectories via MCTS and trains the policy to imitate these trajectories, biasing future exploration toward more promising decisions. In contrast, we estimate the optimal policy using the AWR formulation and train selection heads to directly imitate this policy, eliminating the need for repeated MCTS-based iterations.

\section{Broader context and related work}

\subsection{Step-wise guidance for LLM reasoning}

Different reasoning strategies should be applied as the reasoning state evolves. ArcMemo \cite{ho2025arcmemo} selects modular concepts by prompting the reasoning model, while CoM \cite{jiang2026chain} prompts a meta-agent LLM to select a predefined cognitive mode for the next reasoning step as the reasoning process progresses.

Process reward models (PRMs) may provide reliable evaluations of reasoning trajectories \cite{lightman2023let}. PGTS \cite{li2025policy} trains a graph neural network to select tree-search actions with the goal of maximizing cumulative rewards computed by a PRM or LLM-based evaluation. RLoT \cite{hao2025rl} formulates the reasoning process as a Markov Decision Process (MDP) and trains an RL agent to predict the next logic block based on the current reasoning state, using a PRM as a teacher to guide the training of the RL agent.

In contrast to these approaches, our method enables adaptive step-wise memory selection without requiring an additional PRM to guide training or inference. Moreover, our approach progressively accumulates knowledge as experience grows, rather than restricting the LLM to a fixed set of predefined logic blocks.

\subsection{Self-improvement for LLM reasoning}

Self-improvement methods aim to enhance LLM reasoning performance using the model’s own generations. STaR \cite{zelikman2024star} prompts the LLM to generate reasoning chains and then fine-tunes the model on trajectories that lead to correct answers. B-STaR \cite{zeng2024b} further improves performance by adaptively balancing exploration and exploitation. While effective, these approaches require updating model parameters, which is infeasible for closed-source models.

Self-Refine \cite{madaan2023self} and Reflexion \cite{shinn2023reflexion} instead iteratively refine LLM responses using self-generated feedback. These methods improve performance by modifying the context rather than updating model parameters. However, their effectiveness largely depends on the quality of the generated feedback. BRIDGE \cite{wan2025few} iteratively generates and selects influential demonstrations to enhance the reasoning capability of in-context learning.

Compared to existing approaches, our method constructs and learns to select reusable step-wise memory from confident samples, and then applies this memory to guide reasoning on uncertain samples. Unlike prior self-improvement methods, our approach requires neither model parameter updates nor additional supervision signals or self-feedback. As a result, it is applicable to both open-source and closed-source models in the test-time setting. Moreover, because our method relies on reusable memory derived from previous experience rather than iterative self-feedback, its effectiveness is less dependent on the quality of generated feedback.

\subsection{Retrieval-augmented generation for reasoning} 

Retrieval-augmented generation (RAG) methods dynamically retrieve external information relevant to an input query and condition generation on the retrieved content. ReAct \cite{yao2023react} combines internal and external knowledge by prompting the LLM to search and retrieve information from Wikipedia, aiming to improve reasoning through external knowledge access. HARR \cite{zhang2026reinforcement} formulates multi-hop retrieval as a Markov Decision Process and optimizes the retrieval policy to maximize the F1 score of LLM responses using GRPO \cite{shao2024deepseekmath}. However, existing retrieval-based methods typically operate on a fixed corpus and rely on large-scale training data. In contrast, our work focuses on dynamically expanding memory, where each step-wise instruction is associated with only limited training data.

Rather than retrieving from a fixed corpus, our method incrementally accumulates reusable knowledge by constructing and learning to select step-wise memory. Furthermore, our training framework employs dense step-wise supervision with a one-vs-rest objective, enabling effective learning from limited data while supporting incremental memory expansion during test time.

\subsection{Test-time scaling for LLM reasoning}

Test-time scaling methods improve model performance by allocating additional computation during inference without modifying model parameters. Chain-of-Thought (CoT) prompting \cite{wei2022chain} improves the reasoning ability of large language models (LLMs) by prepending manually crafted reasoning examples to the target question. \cite{jin2024impact} shows that longer reasoning chains consistently improve reasoning performance. Agarwal et al. \cite{agarwal2024many} further demonstrate that increasing the number of few-shot demonstrations mitigates pretraining bias and enhances reasoning ability.

Generating and aggregating multiple responses during inference can further improve reasoning performance. Self-Consistency (SC) \cite{wang2022self} aggregates multiple reasoning trajectories to produce more reliable answers. 

Structured-search methods for reasoning explore multiple step-level continuations and select among them using value or process rewards. Tree-of-Thought (ToT) \cite{yao2023tree} and Graph-of-Thought (GoT) \cite{besta2024graph} introduce structured exploration strategies that improve reasoning performance. rStar \cite{qi2024mutual} explores reasoning states using human-like actions within an MCTS framework and selects trajectories based on mutual consistency. Tree-BoN \cite{qiu2024treebon} reduces the computational cost of tree search by expanding only promising reasoning chains. DORA \cite{wang2025every} adaptively allocates rollout budgets based on PRM scores and trajectory similarity, assigning more computation to reasoning states with lower similarity and higher PRM scores.

However, these methods search over memoryless LLM rollouts without reusable cross-problem knowledge guiding expansion. In contrast, MILES integrates learned, reusable step-level memory into the search, where selection heads choose which sub-instruction to apply at each expansion step.

\section{Detailed method pipeline}

\textbf{Algorithm.} Algorithm~\ref{alg:miles} summarizes one pass through the test stream.

\begin{algorithm}[t]
\caption{MILES: one pass through the test stream}
\label{alg:miles}
\begin{algorithmic}[1]
\Require Frozen LLM with generator $p(\cdot \mid \cdot)$, embedding model, number of initial responses $K$
\State Initialize memory $\mathcal{M} \leftarrow \emptyset$, context codebook $\{\mathbf{q}_{i}\} \leftarrow \emptyset$, selection heads $\{\pi_{\boldsymbol{\theta}_{i}}\} \leftarrow \emptyset$
\For{each test sample $q$ arriving in the stream}
    \State Sample $K$ independent responses from $p$ for $q$
    \If{the $K$ responses are consistent (\emph{confident sample})} \Comment{\emph{confident sample}}
        \State Set pseudo-label $\hat{y} \leftarrow$ majority-vote answer
        \State Run MCTS from $x_{0} = q$ with Layer~1 sampling $\pi_{\text{sim}}$ to expand reasoning states
        \For{each expansion $(x, a_i)$ produced by MCTS}
            \State Roll out from $(x, a_i)$ and set $r_{i}(x) \leftarrow \mathds{1}[\text{rollout answer} = \hat{y}]$
            \State Append $(x, r_{i}(x))$ to per-unit dataset $\mathcal{D}_{i}$
        \EndFor
        \State Summarize new (step-goal, reasoning step) tuples into new memory units
        \State instantiate corresponding selection heads and context codebooks
        \State Update each affected $\pi_{\boldsymbol{\theta}_{i}}$ by minimizing the BCE loss \eqref{eq:cond-loss} on $\mathcal{D}_{i}$
        \State Output the pseudo-label $\hat{y}$ as the answer for $q$
    \Else \Comment{\emph{uncertain sample}}
        \State Run memory-guided tree search using current $(\mathcal{M}, \{\mathbf{q}_{i}\}, \{\pi_{\boldsymbol{\theta}_{i}}\})$ as described above
        \State Output the majority-vote answer over all branches in the response pool
    \EndIf
\EndFor
\end{algorithmic}
\end{algorithm}

\begin{figure}
    \centering
    \includegraphics[width=0.95\linewidth]{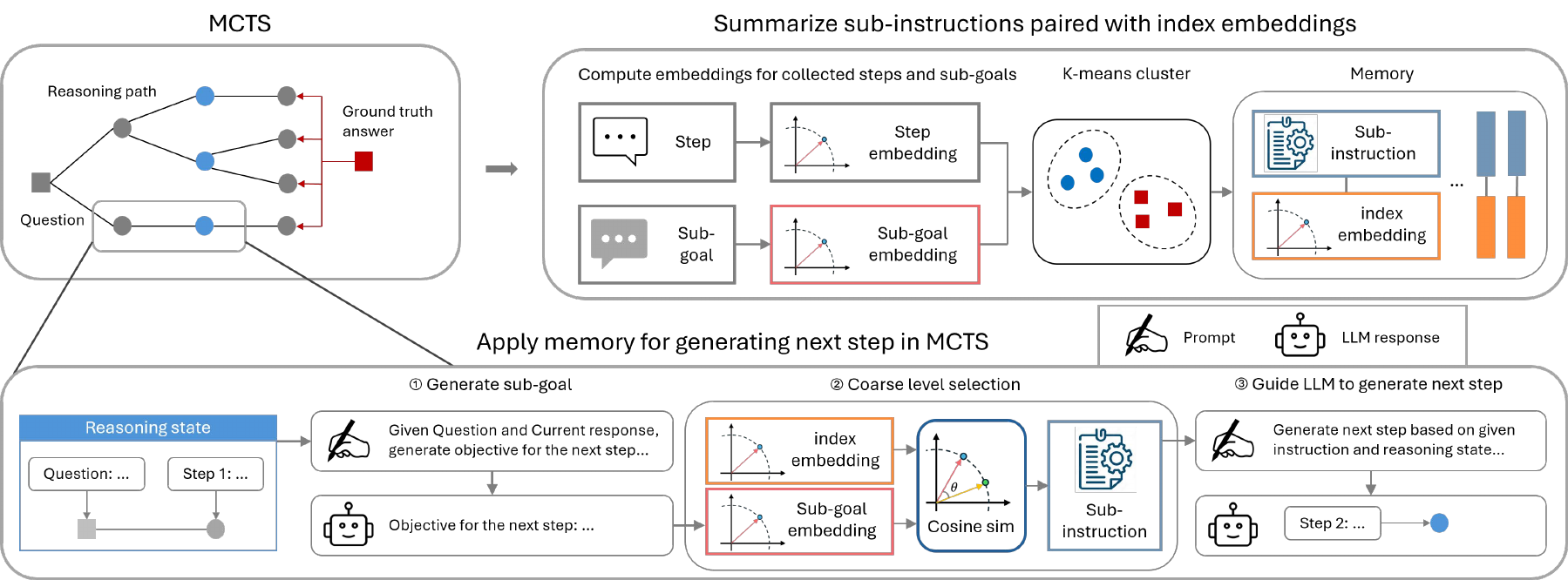}\vspace{-2pt}
    \caption{Overview of MCTS for memory construction and data collection. Sub-instructions are summarized from reasoning trajectories and applied during MCTS to collect training data for the selection heads.} \vspace{-4pt}
    \label{fig:MCTS-proc}
\end{figure}

\subsection{Detailed MCTS pipeline}
\label{sec:mcts-detail}

As discussed in \cite{wang2023math,zhang2024rest}. MCTS enables step-wise evaluation while balancing exploration and exploitation. Following \cite{dou2025enhancing,wu2024beyond,hu-etal-2025-mcts}, our MCTS procedure consists of four main phases: \emph{selection}, \emph{expansion}, \emph{simulation}, and \emph{backpropagation}. Each node represents a reasoning state corresponding to a partial response, with the root node representing the question $q$. Each edge denotes the transition between two reasoning states.\

\textbf{Selection.} Starting from the root node, the MCTS algorithm recursively selects nodes for expansion until reaching a leaf node using the Upper Confidence Bound for Trees (UCT) criterion \cite{kocsis2006bandit}:
\begin{equation}
    \mathrm{UCT}(v)=\frac{Q(v)}{N(v)}+C\cdot \sqrt{\frac{\ln N(\mathrm{parent}(v))}{N(v)}}, \label{eq:uct}
\end{equation}
which balances exploration and exploitation when selecting nodes for expansion. Here, $Q(v)$ denotes the accumulated reward of node $v$, $N(v)$ and $N(\mathrm{parent}(v))$ denote the visit counts of node $v$ and its parent node, respectively, and $C$ is a scalar hyperparameter controlling the exploration–exploitation trade-off. Larger values of $C$ encourage selecting nodes with fewer visits, promoting exploration, while smaller values favor nodes with higher rewards, promoting exploitation. In our setting, the reward is defined as the binary correctness of the final answer, and rewards are backpropagated through all nodes along the reasoning path.\

\textbf{Expansion.} Previous work typically expands a selected node by applying an action from a predefined action space. In contrast, our method generates the next reasoning step by applying a sub-instruction rather than relying on a fixed action space. These instructions are selected based on the similarity between step-goal embeddings and sub-goals. The expansion phase therefore collects application data for sub-instructions, which is subsequently used to train the selection heads.

\textbf{Simulation.} During the simulation phase, the LLM is prompted to complete the reasoning process and produce a final answer, which is then used to evaluate correctness. In our formulation, each node corresponds to a reasoning state. For a reasoning chain with $n$ steps, there are up to $n$ nodes along the reasoning path, where the $i$-th node contains the first $i$ reasoning steps. To prevent excessive rollouts at large depths and maintain diversity in the contexts where sub-instructions are applied, we divide the simulated reasoning path into equal segments and merge multiple steps into a single node, keeping the total search depth fixed.

\textbf{Backpropagation.} After each simulation, the generated answer is evaluated and the leaf node is assigned a binary correctness signal. This signal is then propagated back through all nodes along the reasoning path leading to the final answer, refining the estimated quality of each node and guiding node selection in subsequent search iterations.\

During the initial rollouts from the root node, we do not apply sub-instructions. This stage is therefore equivalent to self-consistency \cite{wang2022self}, and is used to estimate a pseudo ground-truth answer via majority voting. Subsequent backpropagation is then performed based on this estimated answer. Overall, MCTS enables efficient exploration of reasoning trajectories while collecting sub-instruction application data for training the selection heads. The sub-instruction trajectories explored during MCTS are collected as training data for the selection heads. The detailed MCTS procedure is described in Alg.~\ref{alg:mcts}. The details of each step in MCTS are illustrated in Figure~\ref{fig:MCTS-proc}.

\begin{algorithm}[t]
\caption{MCTS for memory construction.}
\label{alg:mcts}
\begin{algorithmic}[1]
\Require Maximum number of rollouts $n_{\text{mcts}}$, number of children nodes $c_{\text{mcts}}$, memory $\mathcal{M}$, question $q$, estimated ground-truth final answer $\hat{y}$.
\State Set $q$ as the root node.
\For{$i=1$ to $n_{\text{mcts}}$}

    \Comment{Selection}
    \State Select a node for expansion using UCT in Eq.~\eqref{eq:uct}.
    \State Extract the reasoning state $x$ from the selected node.
    
    \Comment{Expansion}
    \State Generate a step-goal based on $x$ by prompting the LLM.
    \State Select $c_{\text{mcts}}$ sub-instructions based on the similarity between the step-goal embedding and sub-goals.
    \State Generate the next reasoning step guided by the selected sub-instruction.
    
    \Comment{Simulation}
    \State Prompt the LLM to complete the reasoning process and obtain the final answer $y$.
    
    \Comment{Backpropagation}
    \State Compute the correctness reward $r$, where $r = 1$ if $y = \hat{y}$ and $r = 0$ otherwise.
    \State Backpropagate $r$ through the MCTS trajectory and update the reward and visit count of each node.

    \Comment{Training trajectory collection}
    \State Collect $(x,r)$ for each applied sub-instruction for training selection head.
\EndFor
\end{algorithmic}
\end{algorithm}

\subsection{Detailed memory-guided tree search pipeline}
\label{sec:mgts-detail}

\begin{figure}
    \centering
    \includegraphics[width=0.95\linewidth]{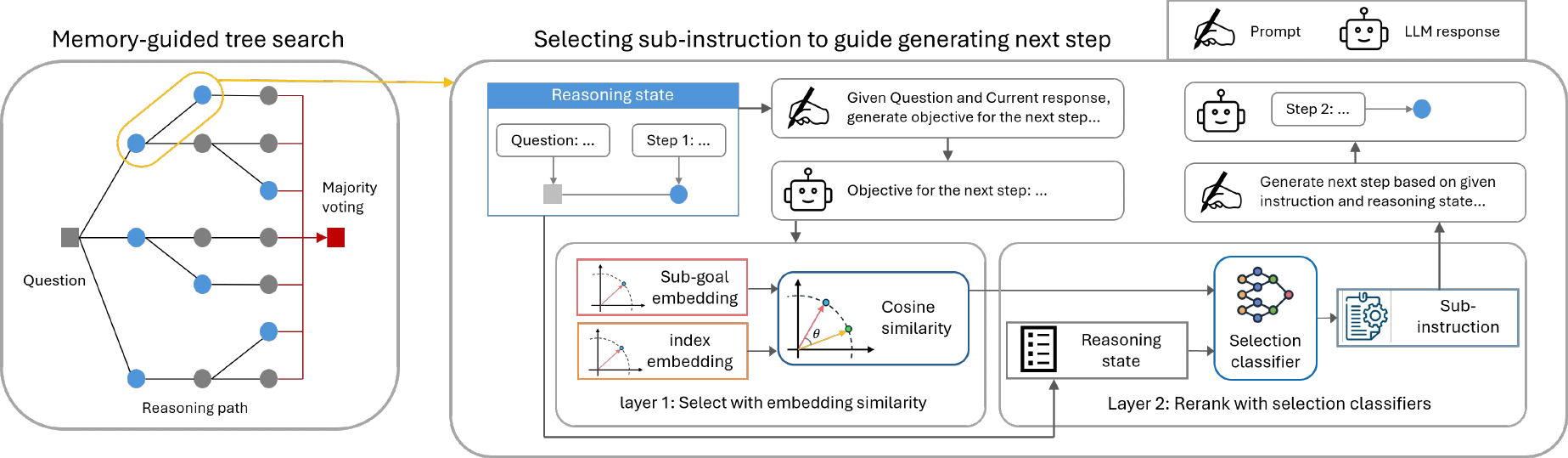}\vspace{-2pt}
    \caption{Overview of memory-guided tree search for inference. The search progressively expands reasoning steps on greater depths. At each step, we select multiple candidate step-wise instructions and rerank them using selection heads. The selected instruction is then used to generate the next reasoning step.} \vspace{-4pt}
    \label{fig:mem-infer}
\end{figure}

When the estimated answers are not consistent on given question, the estimated ground-truth answer and may not be correct and the correctness feedback may be noisy. As discussed in Section~\ref{sec:mcts-detail}, the reward backpropagation and selection in MCTS may bias the expansion towards incorrect answer. Therefore, we do not apply MCTS when model is uncertain. \

During memory-guided tree search, we apply a two-layer selection mechanism to choose sub-instructions for generating the next reasoning step given the current reasoning state. Starting from the question, we iteratively expand reasoning chains over multiple stages while maintaining a response pool that stores all generated trajectories.\

At each stage, we randomly sample a sub-context from the response pool and combine it with the question to form the reasoning state. Based on this state, we select sub-instructions and guide the LLM to generate the next reasoning step. The LLM then completes the reasoning, and the resulting trajectory is added back to the response pool. As more sub-instructions are applied, we progressively increase the length of the sub-context, allowing the reasoning state to incorporate previously guided steps. Similar to MCTS, each node corresponds to a reasoning state and each edge represents a transition between states, with the question serving as the root node. \

Our method predicts the applicability of sub-instructions based on the current reasoning state, but does not evaluate the overall quality of the reasoning state, unlike PRM-based methods \cite{qiu2024treebon,dou2025enhancing,wang2025every}. As a result, we randomly select expansion positions during search. In contrast to TreeBoN \cite{qiu2024treebon}, we do not prune branches, since the quality of a reasoning trajectory cannot be reliably determined. The detailed memory-guided tree search procedure is described in Alg.~\ref{alg:mgts}. Figure~\ref{fig:mem-infer} illustrates the details of each step in memory-guided tree search.

\begin{algorithm}[t]
\caption{Memory-Guided Tree Search}
\label{alg:mgts}
\begin{algorithmic}[1]
\Require Maximum search depth $d_{\text{mgts}}$, number of rollouts per stage $c_{\text{mgts}}$, memory $\mathcal{M}$, question $q$.
\State Initialize the response pool $\mathcal{S}_{\text{res}} \leftarrow \emptyset$.
\For{$i = 1$ to $d_{\text{mgts}}$}

    \Comment{Selection}
    \State Randomly sample $c_{\text{mgts}}$ responses from $\mathcal{S}_{\text{res}}$.
    \State Split the selected responses into reasoning steps.
    \For{each selected response}
        \State Randomly sample a truncation position $p_{\text{step}}$ within 
        $[(i-1)/d_{\text{mgts}} \cdot n_{\text{step}},\ i/d_{\text{mgts}} \cdot n_{\text{step}}]$,
        where $n_{\text{step}}$ denotes the number of reasoning steps.
        \State Use the first $p_{\text{step}}$ reasoning steps as the reasoning state $x$ for expansion.
    \EndFor

    \Comment{Expansion}
    \State Generate multiple step-goals based on $x$ by prompting the LLM.
    \State Select multiple candidate sub-instructions for each step-goal using Layer 1 selection.
    \State Rerank the retrieved sub-instructions using the corresponding selection heads in Layer 2 selection.
    \State Generate the next reasoning step guided by the selected sub-instruction.

    \Comment{Simulation}
    \State Prompt the LLM to complete the reasoning process and obtain the final answer $y$.
    \State Add the generated response to $\mathcal{S}_{\text{res}}$.
\EndFor

\Comment{Answer Estimation}
\State Compute the final answer by majority voting.
\end{algorithmic}
\end{algorithm}

\section{Memory construction details}
\label{sec:mem-constrct}

\subsection{Effective threshold}
\label{sec:eff-range}

The effective threshold is designed to regulate the application of memory unit based on the similarity between step-goal embeddings and sub-goals. We assume that higher similarity between a step-goal embedding and the sub-goal indicates that the corresponding memory unit is more suitable to apply. 

Each memory unit may have a different effective threshold. Intuitively, if a memory unit is summarized from a more dispersed cluster of (step-goal, reasoning step) pairs, it is more general and can be applied under a wider range of conditions. In contrast, instructions derived from tightly clustered pairs are more specialized and should be applied more selectively. Based on this intuition, we compute the effective threshold according to the distribution of the corresponding cluster. 

Specifically, for each memory unit summarized from a cluster of (step-goal, reasoning step) pairs, we first compute the normalized distance between the sub-goal and other step-goal embeddings in the cluster as
\begin{equation}
    d(\mathbf{o}_j,g_i)=0.5\cdot (1-\mathrm{cos}(\mathbf{o}_j,g_i)), \label{eq:emb-dist}
\end{equation}
where $\mathrm{cos}(\cdot,\cdot)$ denotes cosine similarity, $\mathbf{o}_j$ is the $j$-th step-goal embedding in the cluster, and $g_i$ is the sub-goal. We then define the effective threshold as
\begin{equation}
    R_{\text{eff}}=\alpha_{\text{eff}}\cdot \frac{1}{N}\sum_{j=1}^N d(\mathbf{o}_j,g_i),
\end{equation}
where $\alpha_{\text{eff}}$ is a hyperparameter controlling the applicability of the memory unit.

Given the effective threshold defined above, when selecting memory unit given a step-goal embedding, we first compute the distance between the step-goal embedding and the sub-goal using Eq.~\eqref{eq:emb-dist}. If this distance exceeds the effective threshold, we mask the corresponding sub-instruction and exclude it from selection.

\subsection{Incrementally summarize new sub-instructions}

As discussed in Section~\ref{sec:mcts-detail}, our method collects data by retrieving sub-instructions during MCTS. Since each instruction is constrained by an effective threshold, there are cases in which no sub-instruction is selected for the current reasoning state and step-goal. In such cases, we prompt the LLM to generate the next reasoning step that achieves the given step-goal, and store the resulting (step-goal, reasoning step) pair for summarizing new sub-instructions. As sub-instructions are incrementally summarized, their sub-goals gradually cover a larger region of the step-goal embedding space. This expansion enables the memory to handle a broader range of conditions as experience accumulates. The memory construction process is illustrated in detail in Figure~\ref{fig:MCTS-proc}.

\section{Detailed training method}
\label{sec:detail-train}

\begin{algorithm}[t]
\caption{Training selection heads}
\label{alg:train-cls}
\begin{algorithmic}[1]
\Require Training datasets $\mathcal{D}_{1:m}=\{(x,r)\}$ for all sub-instructions, where $m$ is the number of sub-instructions to update, and optimization steps $n_{\text{opt}}$.

\For{$i = 1$ to $m$}

    \Comment{Context Codebook Construction}
    \State Cluster sub-goal embeddings and contextual-condition embeddings using k-means clustering.
    \State Concatenate the centers of all clusters to construct the context codebook.

    \Comment{Training Data Preparation}
    \For{each $x$ in $\mathcal{D}_i$}
        \State Split $x$ into overlapping chunks.
        \State Compute the feature vector using the context codebook according to Eq.~\eqref{eq:model-in-apd}.
    \EndFor
    \State Collect all feature vectors as training data.

    \Comment{selection head Training}
    \State Initialize the selection head for the $i$-th sub-instruction as an MLP.
    \State Train the selection head for $n_{\text{opt}}$ optimization steps using the BCE loss in Eq.~\eqref{eq:bce-loss}.

\EndFor
\end{algorithmic}
\end{algorithm}

The detailed training process is described in Alg.~\ref{alg:train-cls}. 

\paragraph{Construction and application of context code book.} \label{sec:codebook} The step-goal describes the objective the next reasoning step is intended to achieve, while the contextual condition captures the most relevant information for generating that step. Together, they are the key factors determining whether a sub-instruction is applicable to a given reasoning state. During data collection, since each instruction may be applied across multiple reasoning states, we collect the corresponding step-goal and contextual conditions and compute their embeddings. We then cluster these embeddings and use the centroid of each cluster as a context codebook. In this way, context codebooks serve as representative embeddings of similar step-goal and contextual conditions. Each selection head maintains an individual context codebook to identify the information most relevant to its corresponding sub-instruction.

The context codebooks are used to identify relevant information in the reasoning state. Specifically, we first split the reasoning state into overlapping chunks and compute embeddings for each chunk. We then compute the maximum cosine similarity between each context codebook and the chunk embeddings:
\begin{equation}
    \mathbf{m}_i(x) \;=\; [s_{i1}, s_{i2}, \ldots, s_{in}], \qquad
    s_{ij} \;=\; \max_{k \in [1, T]} \mathrm{cos}(\mathbf{q}_{ij}, \mathbf{c}_k),
    \label{eq:model-in-apd}
\end{equation}
where $\mathbf{m}_i(x)$ denotes the feature vector, $\mathbf{q}_{ij}$ and $\mathbf{c}_k$ represent the $j$-th context prototype and the $k$-th chunk embedding, respectively, $T$ is the number of chunks, and $n$ is the number of context prototypes.

Intuitively, the maximum cosine similarity captures whether information relevant to a given context codebook is present in the reasoning state: a higher similarity indicates stronger evidence of its presence. Thus, Eq.~\eqref{eq:model-in-apd} encodes the presence of different step-goal and condition patterns into a fixed-dimensional feature vector, which is then used by the selection head to determine whether a sub-instruction should be applied.

\paragraph{Discussion with Q-learning.} In LLM reasoning tasks, supervision is typically limited to the correctness of the final solution, resulting in sparse and delayed rewards. This makes it challenging to reliably estimate the contribution of individual intermediate actions. Value-based approaches, such as Q-learning, require estimating the expected reward of each action under a given reasoning state, which in practice relies on sufficient rollout samples. When the number of observations per sub-instruction is limited, accurately estimating such values becomes difficult and may suffer from high variance. In our setting, each sub-instruction is applied under diverse reasoning states, and only a small number of samples are available for each instruction–state pair. As a result, value estimation for individual instructions can be unreliable. In extreme cases, if the observed trajectories associated with a given instruction all lead to correct final answers, a value-based model may assign uniformly high values, failing to distinguish the conditions under which the instruction is truly applicable. Instead of explicitly estimating action values, our approach directly approximates closed form optimal policy over instructions using binary correctness signals, avoiding per-action value estimation under limited samples.

\paragraph{Discussion with ColBERT.} ColBERT \cite{khattab2020colbert} measures document relevance to a query by summing the maximum similarity between each context codebook and the document vectors, where both are encoded by a BERT encoder. We use the cluster centers as context codebooks and follow the same idea that identify key information from one sequence by maximum similarity. Different from ColBERT \cite{khattab2020colbert}, we retain the full sequence of maximum similarities between the context codebooks and the document vectors as the input representation, this indicates a combination of different queries.

\paragraph{Method for preventing overfitting.} Since only a limited amount of data is collected for each sub-instruction, the selection heads may overfit to the training data and fail to generalize to new cases. To mitigate this issue, we adopt the following strategies:
\begin{enumerate}
    \item \textbf{Cross-instruction pretraining.} For a target sub-instruction, we first collect data from other instructions with similar sub-goals. A model is then trained on this combined dataset using the binary cross-entropy loss, which encourages the model to learn more diverse patterns for instruction selection. Finally, we fine-tune the classification layer on the training data of the target sub-instruction while keeping the remaining parameters fixed.
    \item \textbf{Input dropout.} During training, we apply dropout to the input dimensions of the selection heads to prevent certain dimensions from dominating the prediction. This regularization encourages the model to rely on more robust features and improves generalization to unseen conditions.
    \item \textbf{Regularization techniques.} In practice, we train multiple selection heads and average their predicted probabilities to obtain a more robust estimate. We additionally employ early stopping and weight decay to further reduce the risk of overfitting to the training data.
\end{enumerate}

\paragraph{Preventing catastrophic forgetting.} During test time, samples arrive sequentially, and the training data for each selection head is incrementally accumulated. As a result, the selection heads must be updated to incorporate newly acquired knowledge. However, updating the model on new data can lead to a significant degradation in performance on previously seen data, a phenomenon known as catastrophic forgetting \cite{mccloskey1989catastrophic}. To mitigate this issue while preventing extra complexity, we adopt a simple rehearsal strategy: we retain all previously collected data and train the selection heads jointly on both new and past data.

\section{Additional experiments}
\label{sec:additional-exp}

\begin{table}[]
\caption{Performance comparison on AIME 2024 and AIME 2025 datasets using GPT-4.1-mini model.}
\label{tab:full-exp}
\centering
\setlength{\tabcolsep}{4pt}
    \begin{tabular}{ll >{\columncolor{ourcol}}c >{\columncolor{ourcol}}c}
    \toprule
    \rowcolor{headcol}
    Method type & Method & AIME 2024 & AIME 2025  \\
    \midrule
    \multirow{2}{*}{Prompt-based methods} & Zero-shot CoT \cite{kojima2022large} & \conf{48.67}{3.02} & \conf{44.67}{2.56} \\
     & AoT \cite{teng2025atom} & \conf{62.00}{3.41} & \conf{49.33}{1.00} \\
     \midrule  
    \multirow{4}{*}{Memory-based methods} & BoT \cite{yang2024buffer} & \conf{52.67}{5.60} & \conf{41.37}{3.62}  \\
     & DC (\textit{w/} coding) \cite{suzgun2025dynamic} & \conf{62.33}{2.26} & \conf{59.33}{3.14} \\
     & DC (\textit{w/o} coding) \cite{suzgun2025dynamic} & \conf{44.00}{2.46} & \conf{31.00}{2.01} \\
    \midrule  
    \multirow{5}{*}{Test-time scaling} & CoT-SC \cite{wang2022self} & \conf{59.67}{1.76} & \conf{53.67}{2.38} \\
     & ToT \cite{yao2023tree} & \conf{54.00}{1.88} & \conf{51.33}{1.23} \\
     & rStar \cite{qi2024mutual} & \conf{58.00}{1.23} & \conf{50.67}{1.88} \\
     & DORA \cite{wang2025every} & \conf{49.33}{1.00} & \conf{29.67}{1.05} \\
    \cmidrule{2-4}
     & Our method & \textbf{\conf{66.36}{1.38}} & \textbf{\conf{60.33}{1.35}} \\
     \bottomrule
    \end{tabular}
\end{table}

To further evaluate the effectiveness of our method, we compare it against a broad range of baselines on the AIME 2024 and AIME 2025 datasets using the GPT-4.1-mini model.  The baselines include:
\textbf{Prompt-based methods} guide LLM reasoning by designing effective prompts.
\begin{itemize}
    \item \textbf{Zero-shot CoT} \cite{kojima2022large} elicits reasoning by simply adding the prefix \textit{“Let’s think step by step.”}.
    \item \textbf{Atom of Thoughts (AoT)} \cite{teng2025atom} models reasoning as a Markovian process, where each state corresponds to a self-contained subproblem.
\end{itemize}
\textbf{Memory-based methods} accumulate reusable knowledge and strategies from prior reasoning experiences and leverage memory to guide LLMs in solving new problems.
\begin{itemize}
    \item \textbf{Buffer of Thoughts (BoT)} \cite{yang2024buffer} distills reasoning templates from past experiences and retrieves the most relevant templates to guide reasoning on new problems.
    \item \textbf{Dynamic CheatSheet (DC)} \cite{suzgun2025dynamic} distills and refines reusable strategies from prior experiences, and applies the most relevant strategy to solve new problems.
\end{itemize}
\textbf{Test-time scaling methods} allocate additional computation during inference to further improve reasoning performance.
\begin{itemize}
    \item \textbf{Self-Consistency (SC)} \cite{wang2022self} aggregates multiple reasoning trajectories to produce more reliable predictions.
    \item \textbf{Tree-of-Thoughts (ToT)} \cite{yao2023tree} introduces a structured reasoning framework that enables LLMs to flexibly explore and expand multiple reasoning states, improving their problem-solving capabilities.
    \item \textbf{rStar} \cite{qi2024mutual} explores reasoning states using human-like actions within a Monte Carlo Tree Search (MCTS) framework, and selects trajectories based on mutual consistency.
    \item \textbf{DORA} \cite{wang2025every} adaptively allocates rollout budgets based on PRM scores and trajectory similarity, assigning more computation to reasoning states that exhibit lower similarity and higher PRM scores.
\end{itemize}

We compare all methods in terms of final accuracy; results are reported in Table~\ref{tab:full-exp}. Compared to other test-time scaling approaches, MILES achieves higher accuracy, demonstrating the effectiveness of its memory construction and selection strategy.

DORA adaptively allocates rollouts to promising reasoning states, but performs worse than other methods, suggesting that PRMs may be unreliable for API-based models. In contrast, MILES improves performance without relying on external PRMs, highlighting its robustness.

Among memory-based methods~\cite{yang2024buffer,suzgun2025dynamic}, DC achieves strong performance when coding is enabled, but its performance degrades substantially without coding. MILES achieves superior performance by leveraging additional computation for training-data collection and memory-guided tree search.

Across all compared methods, MILES consistently outperforms the baselines on both datasets, demonstrating the effectiveness and robustness of our method.

\subsection{Performance with comparable token cost}
\label{sec:comparable-token}

To further evaluate the effectiveness of MILES, we compare it with memory-based baselines under comparable output-token budgets using GPT-4.1-mini. The baselines include Buffer-of-Thoughts (BoT)~\cite{yang2024buffer}, Dynamic CheatSheet (DC)~\cite{suzgun2025dynamic}, ArcMemo (AM)~\cite{ho2025arcmemo}, and ReasoningBank (RB)~\cite{ouyang2025reasoningbank}. DC w/ code and DC w/o code denote Dynamic CheatSheet with and without code execution, respectively. For AIME, we additionally compare with Retrieval-of-Thought (RoT)~\cite{ahmed2025retrieval}. 

All final answers are obtained by majority voting over multiple reasoning rollouts. As the majority-vote results are stable across runs, we omit error bars for this experiment.

\begin{table}[]
\caption{Final-answer accuracy using single-run majority voting under comparable output-token budgets on AIME 2024 and AIME 2025. MILES achieves the highest accuracy on both datasets.}
\label{tab:aime-token}
\centering
    \begin{tabular}{lcccc}
    \toprule
    \multirow{2}{*}{Method} & \multicolumn{2}{c}{AIME 2024} & \multicolumn{2}{c}{AIME 2025} \\ 
    \cmidrule(lr){2-3} \cmidrule(lr){4-5}
     & Acc. & Out tokens & Acc. & Out tokens \\ 
    \midrule
    DC w/ code & 63.33 & 4.10M & \textbf{60.00} & 4.61M \\ 
    DC w/o code & 60.00 & 4.40M & 53.33 & 4.99M \\ 
    BoT & 60.00 & 4.72M & 50.00 & 4.57M \\
    AM  & 50.00 & 3.88M & 46.67 & 3.45M \\
    RoT & 60.00 & 3.99M & 56.67 & 4.59M \\
    RB  & 60.00 & 4.74M & 53.33 & 5.13M \\
    \textbf{Ours} & \textbf{66.67} & 3.99M & \textbf{60.00} & 3.61M \\
    \bottomrule
    \end{tabular}
\end{table}

As shown in Table~\ref{tab:aime-token}, MILES achieves the highest accuracy on both AIME 2024 and AIME 2025. Because all methods use comparable output-token budgets, the performance differences primarily reflect differences in memory design rather than compute. RoT uses external memory constructed from ReasonFlux~\cite{yang2025reasonflux}, which may contribute to its stronger performance relative to the other baselines.

\begin{table}[]
\caption{Final-answer accuracy under comparable output-token budgets on GPQA-Diamond, MMLU-Pro Physics and Engineering subset. MILES achieves the highest accuracy on all the datasets.}
\label{tab:academic-token}
\centering
    \begin{tabular}{lcccccc}
    \toprule
    \multirow{2}{*}{Method} & \multicolumn{2}{c}{GPQA-Diamond} & \multicolumn{2}{c}{MMLU-Pro Physics} & \multicolumn{2}{c}{MMLU-Pro Engineering} \\ 
    \cmidrule(lr){2-3} \cmidrule(lr){4-5} \cmidrule(lr){6-7}
     & Acc. & Out tokens & Acc. & Out tokens & Acc. & Out tokens \\ 
    \midrule
    DC  & 65.66 & 5.69M & 83.50 & 4.54M & 68.00 & 8.72M \\ 
    BoT & 67.17 & 4.81M & 84.00 & 3.48M & 72.50 & 7.54M \\
    AM  & 69.19 & 4.20M & 82.00 & 3.99M & 64.50 & 7.38M \\
    RB  & 66.67 & 5.37M & 82.00 & 4.56M & 64.00 & 7.78M \\
    \textbf{Ours} & \textbf{73.23} & 4.66M & \textbf{85.00} & 3.74M & \textbf{75.00} & 7.46M \\
    \bottomrule
    \end{tabular}
\end{table}

We further evaluate MILES and the baseline methods on three academic benchmarks: GPQA-Diamond, MMLU-Pro Physics, and MMLU-Pro Engineering. As shown in Table~\ref{tab:academic-token}, MILES achieves the highest accuracy across all three benchmarks under comparable output-token budgets, further demonstrating the effectiveness and robustness of our method.

\subsection{Component ablation with more models}

\begin{table}[]
    \centering
    \caption{Ablation study on different components using GPT-4.1-mini and Qwen3-30B-Instruct.}
    \label{tab:ablation-full}
    \begin{tabular}{lcccc}
    \toprule
         Dataset & Layer 1 & Layer 2 & GPT-4.1-mini & Qwen3-30B-Instruct \\
    \midrule
         \multirow{3}{*}{AIME 2024} & \xmark & \xmark & \conf{59.33}{1.00} & \conf{82.67}{1.51} \\
         & \gmark & \xmark & \conf{62.66}{1.01} & \conf{83.00}{1.35} \\
         & \gmark & \gmark & \textbf{\conf{66.36}{1.38}} & \textbf{\conf{85.33}{1.23}} \\
         \midrule
         \multirow{3}{*}{AIME 2025} & \xmark & \xmark & \conf{53.33}{1.59} & \conf{68.34}{1.26} \\
         & \gmark & \xmark & \conf{55.33}{2.01} & \conf{68.67}{2.31} \\
         & \gmark & \gmark & \textbf{\conf{60.33}{1.35}} & \textbf{\conf{72.66}{1.01}} \\
    \bottomrule
    \end{tabular}
\end{table}

In addition to the ablation study in Section~\ref{sec:ablation}, we further evaluate our method across different LLMs, including GPT-4.1-mini and Qwen3-30B-Instruct, on the AIME 2024 and AIME 2025 datasets. Results are reported in Table~\ref{tab:ablation-full}.\

Our method consistently improves performance across different models and datasets, demonstrating its effectiveness and robustness. We also observe that the performance of Qwen3-30B-Instruct is already close to saturation. 

\subsection{Cross-task knowledge transfer}

To further evaluate the knowledge transferability of our memory, we conduct experiments in which memory is constructed and applied on different tasks. For mathematical reasoning, we construct cross-task and in-task memories from MATH-500 and AIME 2020-2024 and apply them to the 13 most uncertain AIME 2025 samples. For academic reasoning, we construct cross-task and in-task memories from the non-Diamond split of GPQA and from MMLU-Pro Physics samples disjoint from the evaluation subset, respectively, and evaluate them on the 25 most uncertain samples from the held-out MMLU-Pro Physics test set.

\begin{table}[]
\caption{Performance on the most uncertain samples from AIME 2025 and MMLU-Pro Physics using memory constructed from different tasks.}
\label{tab:cross-task}
\centering
    \resizebox{\linewidth}{!}{
    \begin{tabular}{lccc}
    \toprule
    Test set & No memory & Cross-task memory & In-task memory \\
    \midrule
    AIME 2025 & \conf{18.46}{2.84} & \conf{35.25}{3.94} (MATH-500) & \textbf{\conf{36.92}{2.32}} (AIME 2020–2024) \\
    MMLU-Pro Physics & \conf{32.00}{4.05} & \conf{36.80}{2.96} (GPQA) & \textbf{\conf{46.00}{2.02}} (MMLU-Pro Physics) \\
    \bottomrule
    \end{tabular}
    }
\end{table}

As shown in Table~\ref{tab:cross-task}, cross-task memory consistently improves over the no-memory setting in both cases, indicating that the stored units capture transferable reasoning strategies rather than only dataset-specific patterns. For the closely related math-to-math transfer, cross-task memory performs comparably to in-task memory. For the more distant graduate-science-to-physics transfer, the improvement is smaller but remains positive, suggesting that transfer gradually decreases as task distance increases rather than failing entirely.

\subsection{Ablation of summarization parameters} 
\label{sec:sum-param}

\begin{table}[t]
\centering

\begin{minipage}[t]{0.58\linewidth}
\centering
\caption{Effect of the minimum cluster size on AIME 2024 with GPT-4.1-mini. Accuracy is stable for moderate cluster sizes ($10$--$15$); cluster sizes that are too small (5) produce overly specific sub-instructions, while too large (20) produce overly generic ones.}
\label{tab:cluster-size}
    \begin{tabular}{cc}
    \toprule
    Cluster size & Accuracy \\
    \midrule
    5 & \conf{64.67}{1.23} \\
    10 & \textbf{\conf{66.36}{1.38}} \\
    15 & \conf{64.00}{1.51} \\
    20 & \conf{60.33}{0.75} \\
    \bottomrule
    \end{tabular}
\end{minipage}
\hfill
\begin{minipage}[t]{0.38\linewidth}
\centering
\caption{Effect of $\alpha_{\text{sum}}$ on AIME 2024 with GPT-4.1-mini. Accuracy is stable within $\alpha_{\text{sum}} \in [1.0, 2.0]$ and drops only when step-goal similarity is ignored ($\alpha_{\text{sum}} = 0$).}
\label{tab:cluster-alpha}
    \begin{tabular}{cc}
    \toprule
    $\alpha_{\text{sum}}$ & Accuracy \\
    \midrule
    0 & \conf{65.00}{1.69} \\
    1.0 & \conf{66.36}{1.38} \\
    2.0 & \textbf{\conf{67.36}{1.00}} \\
    \bottomrule
    \end{tabular}
\end{minipage}

\end{table}

To evaluate the effect of our summarization method, we study two hyperparameters: (1) $\alpha_{\text{sum}}$, which balances the similarities between step-goal embeddings and step embeddings; and (2) the minimum cluster size, which controls the level of abstraction, as only clusters whose size exceeds this threshold are summarized into sub-instructions.
Both are evaluated on AIME 2024 with GPT-4.1-mini in Tables~\ref{tab:cluster-alpha} and~\ref{tab:cluster-size}.

Both hyperparameters show the same robustness pattern. Accuracy remains stable across a moderate range and degrades only at the extremes. For $\alpha_{\text{sum}}$ (Table~\ref{tab:cluster-alpha}), values in $[1.0, 2.0]$ yield stable performance (66.36\% and 67.36\%). When $\alpha_{\text{sum}} = 0$, which entirely removes step-goal similarity from clustering, performance degrades.
For the minimum cluster size (Table~\ref{tab:cluster-size}), values in $[10, 15]$ also remain stable; cluster sizes that are too small (e.g., 5) produce overly specific sub-instructions and drop accuracy, while cluster sizes that are too large (e.g., 20) produce overly generic sub-instructions and also drop. In practice we use $\alpha_{\text{sum}} = 1.0$ and a minimum cluster size in $[10, 20]$ depending on the dataset (Appendix~\ref{sec:exp-detail}).

\begin{table}[t]
\centering

\begin{minipage}[t]{0.48\linewidth}
\centering
\caption{Effect of the effective threshold $\alpha_{\text{eff}}$ on AIME 2025 with GPT-4.1-mini. Accuracy stays above 60.00\% across the moderate range $\alpha_{\text{eff}} \in [2.0, 3.0]$ and only degrades at the extreme values 1.0 and 4.0.}
\label{tab:eff-thd}
    \begin{tabular}{cc}
    \toprule
    Effective threshold & Accuracy \\
    \midrule
    1.0 & \conf{58.34}{1.26} \\
    2.0 & \conf{60.33}{1.35} \\
    3.0 & \textbf{\conf{64.67}{1.23}} \\
    4.0 & \conf{58.34}{1.25} \\
    \bottomrule
    \end{tabular}
\end{minipage}
\hfill
\begin{minipage}[t]{0.43\linewidth}
\centering
\caption{Ablation of the two contributors to the context codebook (contextual conditions and step-goals) on AIME 2024 with GPT-4.1-mini. Both contributors are needed for the best accuracy.}
\label{tab:codebook}
    \begin{tabular}{ccc}
    \toprule
    Contextual & \multirow{2}{*}{Step-goal} & \multirow{2}{*}{Accuracy} \\
    condition &  & \\
    \midrule
    \xmark & \gmark & \conf{62.33}{1.15} \\
    \gmark & \xmark & \conf{64.67}{2.27} \\
    \gmark & \gmark & \textbf{\conf{66.36}{1.38}} \\
    \bottomrule
    \end{tabular}
\end{minipage}

\end{table}

\subsection{Impact of effective threshold}

The effective threshold $\alpha_{\text{eff}}$ controls how broadly a sub-instruction is allowed to be applied: a larger threshold allows more diverse application contexts (and more diverse training trajectories) but raises the risk of applying a sub-instruction to states it does not fit. We evaluate the effect of $\alpha_{\text{eff}} \in \{1.0, 2.0, 3.0, 4.0\}$ on AIME 2025 with GPT-4.1-mini.

Table~\ref{tab:eff-thd} shows that MILES is largely insensitive to $\alpha_{\text{eff}}$ over a wide range: accuracy stays at or above 60.00\% for $\alpha_{\text{eff}} \in \{2.0, 3.0\}$, peaking at 3.0, and only degrades to 58.34\% at the extreme values 1.0 and 4.0, where overly restrictive or broad applicability results in no relevant or additional irrelevant sub-instructions, respectively.
In our experiments we use $\alpha_{\text{eff}} = 2.0$ for math datasets (Appendix~\ref{sec:exp-detail}) without further tuning.

\subsection{Ablation on context codebook}

The context codebook in Layer~2 is built by clustering embeddings of \emph{contextual conditions} and \emph{sub-goals} (Section~\ref{sec:mem-rs}); we ablate the contribution of these two components on AIME 2024 with GPT-4.1-mini. Removing either component drops accuracy, while combining both yields the best result (Table~\ref{tab:codebook}), supporting the assumption that a sub-instruction's applicability is jointly determined by what the next step should achieve i.e., step-goal and what context the current state provides.

\subsection{Ablation on number of candidate sub-instructions}

\begin{wraptable}{R}{0.4\textwidth}
\centering
\caption{Effect of the number of Layer~1 candidates on AIME 2025 with GPT-4.1-mini. Performance degrades when too few candidates are retrieved.}
\label{tab:cand-num}
\begin{tabular}{cc}
\toprule
Candidate number & Accuracy \\
\midrule
8 & \conf{53.33}{1.95} \\
16 & \conf{60.33}{1.35} \\
24 & \textbf{\conf{61.00}{1.15}} \\
\bottomrule
\end{tabular}
\end{wraptable}
In memory-guided tree search, Layer~1 first retrieves candidate sub-instructions based on sub-goal similarity, which are then reranked by Layer~2 selection. We study the effect of the number of Layer~1 candidates on AIME 2025 with GPT-4.1-mini; results are reported in Table~\ref{tab:cand-num}. Performance drops when too few candidates are retrieved (8 candidates), but remains stable with larger candidate sets (16 and 24), indicating that Layer~2 selection effectively identifies useful sub-instructions from the retrieved candidates. In all experiments, we use 16 candidates without tuning this hyperparameter.

\subsection{Ablation on MCTS}

To evaluate the effect of MCTS, We conduct an ablation study on AIME 2024 and AIME 2025 using GPT-4.1-mini, replacing MCTS with random tree search. Specifically, we replace UCT-based selection with random selection during data collection from confident samples, while holding memory construction and selection head training fixed. The accuracy, correct-answer proportion, and instruction-correct rate are reported in Table~\ref{tab:mcts-abl}.

\begin{table}[]
\caption{Performance comparison with and without MCTS on AIME 2024 and AIME 2025 using GPT-4.1-mini.}
\label{tab:mcts-abl}
\centering
\resizebox{\linewidth}{!}{
\begin{tabular}{lcccc}
\toprule
Dataset & Tree search method & Accuracy & Correct-answer proportion	& Instruction-correct rate \\
\midrule
\multirow{2}{*}{\parbox{1.2cm}{AIME\\ 2024}} & MCTS (MILES) & \textbf{\conf{66.36}{1.38}} & \textbf{40.27} & \textbf{55.84} \\
 & Random Tree search & \conf{66.02}{1.01} & 34.21 & 40.82 \\
\midrule
\multirow{2}{*}{\parbox{1.2cm}{AIME\\ 2025}} & MCTS (MILES) & \textbf{\conf{60.33}{1.35}} & \textbf{36.06} & \textbf{45.66} \\
 & Random Tree search & \conf{58.00}{2.30} & 34.18 & 41.35 \\
\bottomrule
\end{tabular}
}
\end{table}

Because only the search procedure varies, this comparison isolates the contribution of MCTS. MILES remains robust when MCTS is replaced by random tree search, indicating that its performance does not depend on MCTS and supporting the memory and selector design as the core contribution. MCTS nevertheless produces higher-quality training data, with a higher correct-answer proportion and instruction-correct rate on both datasets. The improved data quality directly benefits selector training.

\subsection{Effect of selected sub-instructions}
\label{sec:instruct-effect}

To validate that the sub-instructions selected during memory-guided tree search actually help, we compare, under matched reasoning states, the accuracy of reasoning trajectories that include steps guided by a selected sub-instruction against trajectories generated without such guidance. Experiments are conducted on all six benchmarks with GPT-4.1-mini and GPT-4.1; Table~\ref{tab:path-acc} reports the resulting accuracy improvement in percentage points.

\begin{table}[]
\centering
\caption{Accuracy improvement (percentage points) of reasoning paths with vs.\ without applying a selected sub-instruction in memory-guided tree search, under matched reasoning states. Selected sub-instructions consistently improve performance across all six benchmarks and both backbones.}
\label{tab:path-acc}
\begin{tabular}{lcc}
\toprule
\multirow{2}{*}{Dataset} & \multicolumn{2}{c}{Accuracy improvement} \\
\cmidrule{2-3}
 & GPT-4.1-mini & GPT-4.1 \\
\midrule
MATH-500        & 6.48 & 2.23 \\
AIME 2024       & 10.80 & 3.77 \\
AIME 2025       & 9.12 & 2.31 \\
GPQA-Diamond    & 4.36 & 3.21 \\
MMLU-Pro Physics         & 4.12 & 8.73 \\
MMLU-Pro Engineering     & 0.63 & 3.87 \\
\bottomrule
\end{tabular}
\end{table}

As shown in Table~\ref{tab:path-acc}, applying selected sub-instructions consistently improves reasoning performance across all 12 dataset-model cells (range 0.63--10.80 points). This confirms that the sub-instructions retrieved and reranked by our two-layer mechanism are effective at steering the LLM toward correct final answers.

\subsection{Pseudo-label quality}
\label{sec:pseudo-label-quality}

The correctness labels of confident trajectories are determined by comparing the rollout results with the pseudo-labels estimated through self-consistency. To analyze the reliability of these pseudo-labels, we evaluate the pseudo-label accuracy and the proportion of confident samples in experiments using GPT-4.1-mini. We also report self-consistency accuracy as a reference.

\begin{table}[]
\centering
\caption{Pseudo-label accuracy, proportion of samples identified as confident, and self-consistency accuracy.}
\label{tab:pseudo-label}
    \resizebox{\linewidth}{!}{
    \begin{tabular}{lccc}
    \toprule
    Dataset & Pseudo-label acc. & Confident sample proportion (n / total) & Full-set SC acc. \\
    \midrule
    MATH-500        & 96.83 & 442 / 500 & 89.20 \\
    AIME 2024       & 94.44 & 18 / 30 & 59.67 \\
    AIME 2025       & 76.47 & 17 / 30 & 53.67 \\
    GPQA-Diamond    & 79.87 & 159 / 198 & 72.02 \\
    MMLU-Pro Physics         & 88.40 & 181 / 200 & 83.00 \\
    MMLU-Pro Engineering     & 83.03 & 165 / 200 & 72.33 \\
    \bottomrule
    \end{tabular}}
\end{table}

As shown in Table~\ref{tab:pseudo-label}, pseudo-label accuracy exceeds full-set self-consistency accuracy across all benchmarks, indicating that agreement concentrates on more reliable predictions and provides substantially cleaner supervision for memory construction.

To further evaluate the reliability of the pseudo-labels, we conduct an experiment on AIME 2025 using GPT-4.1-mini. We train the selection heads using only truly correct confident samples while holding all other settings fixed. The results are shown in Table~\ref{tab:correct-confident}.

\begin{table}[]
\caption{Performance comparison when selection heads are trained on all confident trajectories or only truly correct confident trajectories. Both settings achieve comparable performance.}
\label{tab:correct-confident}
\centering
\begin{tabular}{lc}
\toprule
Training data & Accuracy \\
\midrule
All confident samples (MILES) & \conf{60.33}{1.35} \\
Truly-correct confident samples only & \conf{60.00}{1.12} \\
\bottomrule
\end{tabular}
\end{table}

Both settings achieve comparable performance and are statistically indistinguishable, indicating that confidently incorrect samples do not adversely affect the learned memory. Training on only truly correct samples slightly reduces variance, while removing memory from confidently incorrect samples slightly lowers accuracy, likely due to reduced coverage. Moreover, because memory is constructed at the step level, an incorrect final answer does not imply that all intermediate steps are incorrect. Our manual analysis (Appendix~\ref{sec:manual-ana}) confirms that confidently incorrect cases retain non-trivial step-level quality.

\subsection{Further robustness analysis}

To further evaluate the robustness of MILES, we conduct experiments by (1) shuffling the sample order on AIME 2024 and AIME 2025 and (2) sampling different subsets from MMLU-Pro Physics and Engineering. We use GPT-4.1-mini as the backbone model. The results are reported in Tables~\ref{tab:aime-order} and~\ref{tab:mmlu-rand-subset}. We note that the sample-order shuffling experiment differs from the sample-order analysis in Section~\ref{sec:ablation}.

\begin{table}[]
\caption{Mean accuracy and 95\% confidence intervals across different sample orders on AIME 2024 and AIME 2025.}
\label{tab:aime-order}
\centering
\begin{tabular}{lc}
\toprule
Dataset & Accuracy \\
\midrule
AIME 2024 & \conf{66.67}{1.95} \\
AIME 2025 & \conf{60.00}{1.59} \\
\bottomrule
\end{tabular}
\end{table}

\begin{table}[]
\caption{Mean accuracy and 95\% confidence intervals across different random subsets from MMLU-Pro Physics and Engineering.}
\label{tab:mmlu-rand-subset}
\centering
\begin{tabular}{lcc}
\toprule
Dataset & Ours & SC \\
\midrule
MMLU-Pro Physics & \conf{84.63}{1.36} & \conf{83.00}{0.65} \\
MMLU-Pro Engineering & \conf{73.75}{1.89} & \conf{72.13}{1.77} \\
\bottomrule
\end{tabular}
\end{table}

The results in both tables demonstrate the robustness of MILES. Across different subsets of MMLU-Pro Physics and Engineering, MILES consistently outperforms the self-consistency (SC) baseline, further supporting the effectiveness of our method.

\subsection{Manual analysis}
\label{sec:manual-ana}

To evaluate the effectiveness of our modular memory, we conduct a manual analysis on the AIME datasets, with raters unaware of sample confidence and final-answer correctness. For each memory-application case, we evaluate the full pipeline—reasoning state $\rightarrow$ generated step-goal $\rightarrow$ selected sub-instruction $\rightarrow$ guided next step $\rightarrow$ resulting trajectory—along six dimensions scored on $\{0,1,2\}$: step-goal validity (whether the step-goal is a reasonable and useful immediate objective), sub-instruction applicability (whether the selected sub-instruction is relevant and applicable to the step-goal), instruction execution (whether the guided step substantively follows the instruction), local correctness, local coherence, and local contribution (whether the guided step makes meaningful progress).

To assess the reliability of our manual analysis, we first score 20 memory-application cases drawn from both confident and uncertain samples across AIME 2024 and AIME 2025, as shown in Table~\ref{tab:manual-test}.
\begin{table}[]
\caption{Evaluation scores across six metrics on AIME 2024 and AIME 2025.}
\label{tab:manual-test}
\centering
\resizebox{\linewidth}{!}{
    \begin{tabular}{lcccc}
    \toprule
    Metric & AIME24 Confident & AIME24 Uncertain & AIME25 Confident & AIME25 Uncertain \\
    \midrule
    Step-goal validity & 1.90 & 0.85 & 1.90 & 0.90 \\
    Sub-instruction applicability & 1.50 & 1.20 & 1.10 & 0.95 \\
    Instruction execution & 1.35 & 1.20 & 1.00 & 0.85 \\
    Local correctness & 1.70 & 1.05 & 1.70 & 1.05 \\
    Local coherence & 1.55 & 1.30 & 1.70 & 1.50 \\
    Local contribution & 1.30 & 1.05 & 1.35 & 1.30 \\
    \bottomrule
    \end{tabular}
}
\end{table}

Cases from confident samples consistently receive higher scores than those from uncertain samples, indicating that these metrics meaningfully capture differences in reasoning quality.

To evaluate the effectiveness of the selection heads, we then score 20 memory-application cases from uncertain samples on AIME 2025, comparing MILES with the variant without selection heads (Exp. 2 in Table~\ref{tab:ablation}). The two methods differ only in their selection mechanism, and the results are shown in Table~\ref{tab:manual-head}.
\begin{table}[]
\caption{Evaluation scores across six metrics with and without selection heads on AIME 2025.}
\label{tab:manual-head}
\centering
    \begin{tabular}{lcc}
    \toprule
    Metric & Baseline (no heads) & MILES \\
    \midrule
    Step-goal validity & 0.60 & \textbf{0.90} \\
    Sub-instruction applicability & 0.85 & \textbf{0.95} \\
    Instruction execution & 0.85 & 0.85 \\
    Local correctness & 1.00 & \textbf{1.05} \\
    Local coherence & 1.25 & \textbf{1.50} \\
    Local contribution & 0.80 & \textbf{1.30} \\
    \bottomrule
    \end{tabular}
\end{table}

MILES improves the dimensions most directly related to effective reasoning composition—local contribution (1.30 vs. 0.80) and local coherence (1.50 vs. 1.25)—indicating that the selected guidance promotes meaningful and coherent progress rather than superficial contextual matching. We also find that step-goals associated with sub-instructions selected by the learned heads are of higher quality, providing further evidence that MILES selects guidance aligned with reasonable intermediate objectives rather than producing correct answers through unsound reasoning.

Since reasoning is a multi-step process, we further compare confidently correct and confidently incorrect samples on AIME 2025, as shown in Table~\ref{tab:manual-conf}.
\begin{table}[]
\caption{Evaluation scores across six metrics for confidently correct and confidently incorrect samples on AIME 2025.}
\label{tab:manual-conf}
\centering
    \begin{tabular}{lcc}
    \toprule
    Metric & Correct & Wrong \\
    \midrule
    Step-goal validity & 1.85 & 1.35 \\
    Sub-instruction applicability & 1.30 & 1.05 \\
    Instruction execution & 1.05 & 0.90 \\
    Local correctness & 1.85 & 1.40 \\
    Local coherence & 1.60 & 1.45 \\
    Local contribution & 1.50 & 1.20 \\
    \bottomrule
    \end{tabular}
\end{table}

Confidently correct cases score higher across all six dimensions, indicating that correct final answers are associated with consistently stronger intermediate reasoning. At the same time, confidently incorrect cases retain non-trivial local quality, supporting the step-level robustness of MILES: an incorrect final answer does not imply that all intermediate steps are invalid. We note that sub-instructions are intentionally formulated as high-level, generalizable guidance (e.g., ``rewrite or simplify through substitution or factorization"), so their direct applicability may be moderate even when they provide useful guidance.

\section{Efficiency discussion}

\subsection{Reducing time delay via parallel data collection.} 

Our method relies on MCTS with multiple rollouts to collect teacher trajectories for training the selection heads, which can be time-consuming. To mitigate this overhead, we leverage parallel computation to reduce data collection latency. \

Specifically, after the initial rollouts in MCTS, we estimate a pseudo ground-truth answer via majority voting and check for consistency across rollouts. If the answers are consistent, the LLM can directly return the final prediction, as additional data collection does not affect the outcome. In this case, trajectory collection can proceed asynchronously in the background, while the system handles new queries in parallel. If the answers are inconsistent, we instead invoke memory-guided tree search to continue reasoning and determine the final answer.

\subsection{Applying KV-cache to reduce computation.} 

During both MCTS and memory-guided tree search, new branches are expanded from existing reasoning steps by prompting the LLM with the question, previous steps, and a selected sub-instruction to generate the next reasoning step. For open-source LLMs, the KV cache of previously processed tokens can be reused. As a result, the model avoids recomputing representations for earlier tokens, significantly reducing the computational cost of tree search.

\section{Broader impact}
\label{sec:broader-impact}


We propose a framework that learns to select sub-instructions at test time to improve reasoning performance. Our method does not require curated training datasets or process reward models (PRMs), enabling effective learning under limited data. Moreover, the memory construction process incrementally incorporates new sub-instructions as experience accumulates, allowing the system to adapt to diverse scenarios by updating memory rather than modifying LLM parameters.\

As our approach relies on textual memory that is compatible across different LLMs, it enables knowledge transfer across models. In particular, trajectories collected from stronger models can be used to train selection policies for weaker models, potentially improving accessibility to high-quality reasoning capabilities without requiring large-scale training resources. Importantly, the teacher model does not need to be uniformly stronger across all domains. In specific areas, a smaller but more specialized model can provide high-quality trajectories to train selection policies for other models, enabling flexible and domain-specific knowledge transfer.\

However, since the selection policy is learned from collected trajectories, it may inherit biases or errors present in the data, potentially leading to misleading or suboptimal outputs. Additionally, the accumulation of memory over time may amplify such biases if not properly controlled. To mitigate these risks, the use of textual memory allows for transparent inspection, as well as LLM-based verification and feedback. Future work may explore mechanisms for monitoring, filtering, and refining memory and selection policies through self-checking or external supervision.

\section{Implementation details}
\label{sec:exp-detail}

\subsection{MCTS details}

The detailed hyperparameters of MCTS are reported in Table~\ref{tab:mcts-detail}. For more complex datasets, such as AIME 2024 and AIME 2025, we use greater search depth and more rollouts to collect training trajectories, while keeping the same hyperparameters for other datasets. The exploration constant $C$ in Eq.~\eqref{eq:uct} is set to 1.0 across all experiments. 

\begin{table}[]
\caption{Detailed MCTS hyperparameters for experiments.}
\label{tab:mcts-detail}
\centering
\begin{tabular}{lcccc}
\toprule
Dataset & Rollouts & Depth & Children Nodes & $C$ \\
\midrule
MATH-500 & 20 & 4 & 3 & 1.0 \\
AIME 2024 & 40 & 5 & 3 & 1.0 \\
AIME 2025 & 40 & 5 & 3 & 1.0 \\
GPQA-Diamond & 20 & 4 & 3 & 1.0 \\
MMLU-Pro Physics & 20 & 4 & 3 & 1.0 \\
MMLU-Pro Engineering & 20 & 4 & 3 & 1.0 \\
\bottomrule
\end{tabular}
\end{table}

\subsection{Memory construction details}

When clustering (step, step-goal) embeddings, instead of fixing the number of clusters in k-means, we impose a minimum cluster size to ensure that sub-instructions are summarized from sufficiently diverse reasoning steps, resulting in more reusable abstractions. Specifically, we set the minimum cluster size in the range of $[10, 20]$, depending on the similarity among queries within each dataset.

For math tasks, we set the $\alpha_{\text{eff}}$ to 2.0, as sub-instructions tend to generalize well across queries. In contrast, for academic tasks, we use a lower threshold in the range of $[1.0, 1.5]$, since sub-instructions are less transferable across different queries.

\subsection{Training details}

We use a 3-layer MLP with 16 hidden units as the selection head. For each sub-instruction, we train 10 independent classifiers and average their predicted application probabilities to improve robustness. We periodically update the classifiers during test-time processing: after every 2 questions, and whenever the number of training trajectories increases by 20. For smaller datasets such as AIME 2024 and AIME 2025, we instead update the selection heads after each question.

The context prototypes correspond to the cluster centers obtained from embeddings of contextual conditions and step-goals. We set the number of context prototypes (i.e., the number of clusters in k-means clustering) to $\min(N^{\text{app}}_i \cdot 0.2, 512)$, where $N^{\text{app}}_i$ denotes the number of application cases for the $i$-th sub-instruction. The number of context prototypes increases gradually as the application cases become more diverse. We cap the maximum number at 512 to avoid overly complex context codebooks while maintaining clustering efficiency.

\subsection{Memory guided tree search details}

In our method, the application of memory-guided tree search is determined by the majority-answer proportion, i.e., answer consistency. Since different models exhibit different levels of answer consistency, we set the consistency threshold slightly lower than the zero-shot CoT accuracy measured on a small validation subset. We set the number of routing rollouts to 10 for all experiments. The detailed hyperparameters of memory-guided tree search are reported in Table~\ref{tab:mgts-detail}. For more complex datasets, such as AIME 2024 and AIME 2025, we use greater search depths and more rollouts to enable broader application of sub-instructions during reasoning.\

The parameter $\beta_{\text{c2f}}$ in Eq.~\eqref{eq:combined-score} controls the balance between selection and semantic relevance. We set $\beta_{\text{c2f}}=1.0$ for math tasks, so that selection is determined solely by the selection heads, and use values in the range $[0.4, 0.6]$ for academic tasks to place greater weight on semantic relevance.

\begin{table}[]
\caption{Detailed memory-guided tree search hyperparameters for experiments.}
\label{tab:mgts-detail}
\centering
\begin{tabular}{lcccc}
\toprule
Dataset & Stage & Rollout per stage & Candidate instructions & $\beta_{\text{c2f}}$ \\
\midrule
MATH-500 & 5 & 4 & 16 & 1.0 \\
AIME 2024 & 6 & 4 & 16 & 1.0 \\
AIME 2025 & 6 & 4 & 16 & 1.0 \\
GPQA-Diamond & 5 & 4 & 16 & 0.4-0.6 \\
MMLU-Pro Physics & 5 & 4 & 16 & 0.4-0.6 \\
MMLU-Pro Engineering & 5 & 4 & 16 & 0.4-0.6 \\
\bottomrule
\end{tabular}
\end{table}

\subsection{Decoding details}

For all experiments using GPT-4.1-mini and GPT-4.1, we use the OpenAI API service. The decoding temperature is set to 0.0 for all experiments. The maximum response length is determined according to task complexity, ranging from 2048 to 16384 tokens. For experiments with GPT-4.1-mini, we use the \texttt{GPT-4.1-mini-2025-04-14} checkpoint.

For experiments using open-source models, we employ the vLLM framework \cite{kwon2023efficient} for local decoding and the OpenRouter API for online decoding. For all Qwen3 models, we follow the decoding settings described in the original paper \cite{yang2025qwen3}. For the GPT-OSS-20B model, we set the decoding temperature to 1.0 for all experiments.

All experiments use \texttt{text-embedding-3-small} from the OpenAI API service for embedding computation. The embedding dimension is fixed at 1536 across all experiments.

\subsection{Hyperparameter tuning}

We divide the hyperparameters into two groups: memory construction and memory application. We then examine three stages of the reasoning process: (1) step-goal generation, (2) two-layer selection, and (3) rollout and correctness evaluation. Based on these observations, we identify which components contribute to errors and adjust the corresponding hyperparameters. Specifically, if a suitable memory unit exists but the two-layer selection mechanism selects an inappropriate one, we tune the selection hyperparameters. If no suitable memory unit is available, we consider improving the diversity and generalizability of memory units or adjusting the selection criteria to allow fallback to the zero-shot baseline.

\section{Computation resources and assets}
\label{sec:resources}

For experiments using API-based models, we train the selection heads on a single RTX 3090 GPU and access the models via LiteLLM \url{https://docs.litellm.ai/}. For open-source models, we either (1) deploy the models locally on a server with 8 A100 GPUs, or (2) access inference services through OpenRouter \url{https://openrouter.ai/}.

Open-sourced model URLs:
\begin{itemize}
    \item Qwen3-30B-Instruct: \url{https://huggingface.co/Qwen/Qwen3-30B-A3B-Instruct-2507}.
    \item Qwen3-4B-Instruct: \url{https://huggingface.co/Qwen/Qwen3-4B-Instruct-2507}.
    \item GPT-OSS-20B: \url{https://huggingface.co/openai/gpt-oss-20b}.
\end{itemize}

Dataset URLs:
\begin{itemize}
    \item AIME datasets: \url{https://artofproblemsolving.com/wiki/index.php/AIME_Problems_and_Solutions}.
    \item MATH-500: \url{https://huggingface.co/datasets/HuggingFaceH4/MATH-500}.
    \item GPQA-Diamond: \url{https://huggingface.co/datasets/fingertap/GPQA-Diamond}.
    \item MMLU-Pro: \url{https://huggingface.co/datasets/TIGER-Lab/MMLU-Pro}.
\end{itemize}

\section{Case study}

\definecolor{instruct_color}{HTML}{dc5888}
\definecolor{common_color}{HTML}{7a70a7}

To illustrate the effectiveness of sub-instructions, we present two case studies from the algebra and geometry categories of MATH-500 during memory-guided tree search. For each case, we first show the response generated by Zero-shot CoT, followed by the response obtained by applying a selected sub-instruction at an intermediate reasoning step, together with the corresponding sub-instruction. Selected sub-instructions are highlighted in \textcolor{instruct_color}{red}, while shared responses between the two reasoning trajectories are highlighted in \textcolor{common_color}{purple}.

\subsection{Algebra case}

\begin{tcolorbox}[
    breakable,
    colback=white,
    colframe=blue!50,
    title=Case 1: Algebra,
    fonttitle=\bfseries,
    boxrule=0.5pt
]

\textbf{Question:} Let $x,$ $y,$ and $z$ be positive real numbers such that $xyz = 2.$  Find the minimum value of $x^4 + 4y^2 + 4z^4$. \\
\tcblower

\textbf{Zero-shot CoT:}\\

\textcolor{common_color}{Step 1: We can apply the AM-GM inequality to the terms $x^4$, $2y^2$, $2y^2$, $2z^4$, and $2z^4$ to relate their sum to their product, taking into account the constraint $xyz = 2$.} \\

Step 2: Let's rewrite the expression as follows:
\begin{equation}
    x^4 + 4y^2 + 4z^4 = x^4 + 2y^2 + 2y^2 + 2z^4 + 2z^4 \notag
\end{equation}
Now, we have five terms: $x^4$, $2y^2$, $2y^2$, $2z^4$, $2z^4$. \\

Step 2: By the AM-GM inequality, we have
\begin{equation}
    x^4 + 2y^2 + 2y^2 + 2z^4 + 2z^4 \geq 5 \left(x^4 \cdot (2y^2)^2 \cdot (2z^4)^2\right)^{1/5}. \notag
\end{equation} \\

Step 3: Let's simplify the product inside the root:
\begin{equation}
    x^4 \cdot (2y^2)^2 \cdot (2z^4)^2 = x^4 \cdot 4y^4 \cdot 4z^8 = 16 x^4 y^4 z^8. \notag
\end{equation} \\

Step 4: Recall the constraint $xyz = 2$. We can write $z = \frac{2}{xy}$, so
\begin{equation}
    z^8 = \left(\frac{2}{xy}\right)^8 = \frac{256}{x^8 y^8}. \notag
\end{equation}
Thus,
\begin{equation}
    16 x^4 y^4 z^8 = 16 x^4 y^4 \cdot \frac{256}{x^8 y^8} = 16 \cdot 256 \cdot x^4 y^4 / (x^8 y^8) = 4096 \cdot \frac{1}{x^4 y^4}. \notag
\end{equation}\\

Step 5: Substitute this back into the AM-GM inequality:
\begin{equation}
    x^4 + 4y^2 + 4z^4 \geq 5 \left(4096 \cdot \frac{1}{x^4 y^4}\right)^{1/5} = 5 \cdot 4096^{1/5} \cdot (x^4 y^4)^{-1/5}. \notag
\end{equation}\\

Step 6: Compute $4096^{1/5}$. Since $4096 = 2^{12}$, $4096^{1/5} = 2^{12/5} = 2^{2.4}$.\\

Step 7: Now, let's try to achieve equality in AM-GM, which occurs when all terms are equal:
\begin{equation}
    x^4 = 2y^2 = 2y^2 = 2z^4 = 2z^4. \notag
\end{equation}
So $x^4 = 2y^2 = 2z^4$. \\

Step 8: Set $x^4 = 2y^2$ and $x^4 = 2z^4$. Then $y^2 = \frac{x^4}{2}$ and $z^4 = \frac{x^4}{2}$, so $z^2 = \left(\frac{x^4}{2}\right)^{1/2} = \frac{x^2}{\sqrt{2}}$.

Step 9: Now, use the constraint $xyz = 2$:
\begin{equation}
    x \cdot y \cdot z = 2. \notag
\end{equation}
Recall $y = \frac{x^2}{\sqrt{2}}$ and $z = \frac{x^2}{\sqrt{2}}$, so
\begin{equation}
    x \cdot \frac{x^2}{\sqrt{2}} \cdot \frac{x^2}{\sqrt{2}} = 2 \implies x \cdot \frac{x^4}{2} = 2 \implies \frac{x^5}{2} = 2 \implies x^5 = 4 \implies x = 4^{1/5} = 2^{2/5}. \notag
\end{equation}\\

Step 10: Now, $y = \frac{x^2}{\sqrt{2}} = \frac{(2^{2/5})^2}{\sqrt{2}} = \frac{2^{4/5}}{2^{1/2}} = 2^{4/5 - 1/2} = 2^{(8-5)/10} = 2^{3/10}$. 

Similarly, $z = 2^{3/10}$.\\

Step 11: Now, compute the minimum value:
\begin{equation}
    x^4 + 4y^2 + 4z^4 = (2^{2/5})^4 + 4(2^{3/10})^2 + 4(2^{3/10})^4. \notag
\end{equation}

Calculate each term:
- $x^4 = (2^{2/5})^4 = 2^{8/5}$
- $y^2 = (2^{3/10})^2 = 2^{6/10} = 2^{3/5}$
- $z^4 = (2^{3/10})^4 = 2^{12/10} = 2^{6/5}$

So,
\begin{equation}
    x^4 + 4y^2 + 4z^4 = 2^{8/5} + 4 \cdot 2^{3/5} + 4 \cdot 2^{6/5} \notag
\end{equation}

Step 12: This is the minimum value. Thus, the answer is $\boxed{2^{8/5} + 4 \cdot 2^{3/5} + 4 \cdot 2^{6/5}}$\\

\medskip
\hrule
\bigskip

\textbf{Apply sub-instruction}\\

<Sub-instruction>

\textcolor{instruct_color}{Rewrite or simplify algebraic expressions or equations by substituting variables, factoring, or using properties of roots and exponents to transform them into more manageable forms.}

</Sub-instruction> \\

Step 2: Substitute $z = \frac{2}{xy}$ into the expression to rewrite it as $x^4 + 4y^2 + 4\left(\frac{2}{xy}\right)^4$. \\

Step 3: Simplify the expression:
\begin{equation}
    x^4 + 4y^2 + 4\left(\frac{2}{xy}\right)^4 = x^4 + 4y^2 + 4 \cdot \frac{16}{x^4y^4} = x^4 + 4y^2 + \frac{64}{x^4y^4} \notag
\end{equation} \\

Step 4: Let $a = x^4$ and $b = y^2$, so $x = a^{1/4}$, $y = b^{1/2}$, and $z = \frac{2}{xy} = \frac{2}{a^{1/4}b^{1/2}}$. \\

Step 5: Substitute these into the constraint $xyz = 2$:
\begin{equation}
    x y z = a^{1/4} b^{1/2} \cdot \frac{2}{a^{1/4} b^{1/2}} = 2 \notag
\end{equation}
So the constraint is satisfied for all positive $a, b$.\\

Step 6: The expression to minimize is:
\begin{equation}
    a + 4b + \frac{64}{ab^2} \notag
\end{equation} \\

Step 7: Let’s use symmetry and try $a = 2b$ (since the terms $a$ and $b$ appear with different powers, but the last term is symmetric in $a$ and $b^2$). Set $a = k b$, so:
\begin{equation}
    a + 4b + \frac{64}{ab^2} = k b + 4b + \frac{64}{k b^3} \notag
\end{equation}
Let’s try $k = 2$:
\begin{equation}
    2b + 4b + \frac{64}{2b^3} = 6b + \frac{32}{b^3} \notag
\end{equation}
Now, minimize $6b + \frac{32}{b^3}$ for $b > 0$. \\

Step 8: Take the derivative with respect to $b$ and set to zero:
\begin{equation}
    \frac{d}{db}\left(6b + \frac{32}{b^3}\right) = 6 - \frac{96}{b^4} = 0 \implies 6 = \frac{96}{b^4} \implies b^4 = 16 \implies b = 2 \notag
\end{equation}\\

Step 9: Substitute $b = 2$ back:
\begin{equation}
    6b + \frac{32}{b^3} = 6 \times 2 + \frac{32}{8} = 12 + 4 = 16 \notag
\end{equation}\\

Step 10: Recall $a = 2b = 4$, so $x^4 = 4 \implies x = \sqrt{2}$, $y^2 = 2 \implies y = \sqrt{2}$, and $z = \frac{2}{xy} = \frac{2}{2} = 1$. \\

Step 11: Check the constraint: $xyz = \sqrt{2} \cdot \sqrt{2} \cdot 1 = 2$, which is satisfied. \\

Step 12: Compute the original expression:
\begin{equation}
    x^4 + 4y^2 + 4z^4 = 4 + 4 \times 2 + 4 \times 1 = 4 + 8 + 4 = 16 \notag
\end{equation}
Final answer: \boxed{16}

\end{tcolorbox}

In the Zero-shot CoT response, the LLM attempts to solve the problem using the AM-GM inequality. However, this formulation leads to complicated computations and ultimately produces an incorrect answer. In contrast, the selected sub-instruction guides the LLM to first transform the expression into a more manageable form. As a result, the LLM simplifies the formula and successfully reaches the correct answer.

\subsection{Geometry case}

\begin{tcolorbox}[
    breakable,
    colback=white,
    colframe=blue!50,
    title=Case 2: Geometry,
    fonttitle=\bfseries,
    boxrule=0.5pt
]

\textbf{Question:} Quadrilateral $ABCD$ is a square with area 16 square inches. The figure represents the pieces of a Chinese tangram in which all the triangles are isosceles and piece "e'' is a square. What is the area of the gray piece, in square inches?

\begin{tikzpicture}[scale=1]

\fill[gray!70] (2,0) -- (4,2) -- (4,0) -- cycle;

\draw[thick] (0,0) rectangle (4,4);

\draw[thick] (0,0) -- (4,4);

\draw[thick] (1,1) -- (2,0) -- (4,2);
\draw[thick] (0,4) -- (3,1) -- (3,3);

\node[left]  at (0,4) {$A$};
\node[right] at (4,4) {$B$};
\node[right] at (4,0) {$C$};
\node[left]  at (0,0) {$D$};

\node at (2,1) {$e$};

\end{tikzpicture}\\
\tcblower

\textbf{Zero-shot CoT:}\\

\textcolor{common_color}{Step 1: Since the area of square $ABCD$ is $16$ square inches, its side length is $\sqrt{16} = 4$ inches.}\\

\textcolor{common_color}{Step 2: The tangram divides the square into 7 pieces: 5 triangles, 1 parallelogram, and 1 small square (piece "e"). All triangles are isosceles right triangles, and piece "e" is a square.} \\

Step 3: The gray piece in the diagram is the small triangle at the bottom right, adjacent to $C$ and below piece "e". We need to find its area. \\

Step 4: Let's analyze the construction. The classic tangram divides the square into the following pieces (by area, as fractions of the whole square):

- 2 large triangles (each $\frac{1}{4}$ of the area)

- 1 medium triangle ($\frac{1}{8}$)

- 2 small triangles (each $\frac{1}{16}$)

- 1 square ($\frac{1}{8}$)

- 1 parallelogram ($\frac{1}{8}$) \\

Step 5: The gray triangle is one of the two smallest triangles, each with area $\frac{1}{16}$ of the total area.\\

Step 6: Since the total area is 16 square inches, the area of the gray piece is $16 \times \frac{1}{16} = 1$ square inch. \\

Final answer: \boxed{1}

\medskip
\hrule
\bigskip

\textbf{Apply sub-instruction}\\

<Sub-instruction>

\textcolor{instruct_color}{Calculate the area of a triangle with sides $a$, $b$, and $c$ by first finding the semi-perimeter $s = \frac{a+b+c}{2}$, then applying Heron's formula: $A = \sqrt{s(s-a)(s-b)(s-c)}$.}

</Sub-instruction> \\

Step 3: To find the area of the gray triangle with vertices at $(2,0)$, $(4,0)$, and $(4,2)$, first compute the side lengths: $a = 2$ (from $(2,0)$ to $(4,0)$), $b = 2$ (from $(4,0)$ to $(4,2)$), and $c = \sqrt{(4-2)^2 + (2-0)^2} = \sqrt{4+4} = 2\sqrt{2}$ (from $(2,0)$ to $(4,2)$). \\

Step 4: The triangle is a right triangle with legs of length $2$ (from $(2,0)$ to $(4,0)$ and from $(4,0)$ to $(4,2)$), so its area is $\frac{1}{2} \times 2 \times 2 = 2$ square inches. \\

Step 5: Therefore, the area of the gray piece is \boxed{2} square inches.

\end{tcolorbox}

In the Zero-shot CoT response, the LLM directly estimates the area of the gray region based on the partition of the square. However, it identifies the wrong triangle, which leads to an incorrect final answer. In contrast, the selected sub-instruction guides the LLM to compute the area of the gray region using Heron’s formula. Consequently, the LLM first computes the lengths of the edges of the gray region in the following step. Although the LLM does not explicitly apply Heron’s formula afterward, the computed edge lengths help it derive the correct area.

From the above cases, we observe that the selected sub-instructions guide the LLM toward more promising reasoning directions, ultimately leading to correct final answers. These results suggest that our method effectively achieves the objective of selecting appropriate sub-instructions to maximize the probability of generating correct solutions, demonstrating the effectiveness of our approach.

\section{All prompts}

To ensure generalizability across tasks, we use a unified set of prompts for all experiments. To maintain compatibility with smaller models such as Qwen3-30B-Instruct and GPT-OSS-20B, we keep the prompts simple and straightforward.\

\textbf{Generate step-goal.} This prompt generates a step-goal for the next reasoning step based on the current reasoning state. The resulting step-goal is used to select a sub-instruction by computing cosine similarity between its embedding and the stored sub-goals. The selected sub-instruction then guides the generation of the next reasoning step. 

\begin{tcolorbox}[
    breakable,
    colback=gray!10,
    colframe=gray!50,
    title=Generate step-goal,
    fonttitle=\bfseries,
    boxrule=0.5pt
]
\textbf{Instruction:} Given the Question and Reasoning so far, output the immediate next-step objective that directly contributes toward achieving the final goal based on current progress.
Respond in one sentence only. \\

\textbf{Question:}

<question>\\

\textbf{Reasoning so far:}

<reasoning so far>
\end{tcolorbox}
\

\textbf{Generate next step with sub-instruction.} This prompt guides the LLM to generate the next reasoning step conditioned on the selected sub-instruction and the current reasoning state. In both MCTS and memory-guided tree search, generating the next reasoning step corresponds to expanding a new branch from the existing reasoning state.

\begin{tcolorbox}[
    breakable,
    colback=gray!10,
    colframe=gray!50,
    title=Generate next step with sub-instruction,
    fonttitle=\bfseries,
    boxrule=0.5pt
]
\textbf{Instruction:} Apply the given instruction to generate the NEXT reasoning step only.

- Output a single step, one sentence.

- Be precise. Use exact numbers, variables, or terms from the question and prior steps.

- Do not restate irrelevant details.

- Keep it in the same style as the previous steps.

- If the reasoning is complete, present the final answer as (\textbackslash boxed\{X\}), where X is the correct answer. \\

\textbf{Question:}

<question>\\

\textbf{Reasoning so far:}

<reasoning so far>\\

\textbf{Relevant instruction:}

<relevant instruction>

\end{tcolorbox}
\

\textbf{Generate next step without sub-instruction.} When no sub-instruction is selected, we prompt the LLM to generate the next reasoning step to achieve the given step-goal. The resulting (step-goal, reasoning step) pair is then stored for summarizing new sub-instructions.

\begin{tcolorbox}[
    breakable,
    colback=gray!10,
    colframe=gray!50,
    title=Generate next step without sub-instruction,
    fonttitle=\bfseries,
    boxrule=0.5pt
]
\textbf{Instruction:} Generate the Next reasoning step that best fulfills the given Objective of the next step in the context of the Question and Reasoning so far.

- Output a single step, one sentence.

- Be precise. Use exact numbers, variables, or terms from the question and prior steps.

- Do not restate irrelevant details.

- Keep it in the same style as the previous steps.

- If the reasoning is complete, present the final answer choice as (\textbackslash boxed\{X\}), where X is the correct answer. \\

\textbf{Question:}

<question>\\

\textbf{Reasoning so far:}

<reasoning so far>\\

\textbf{Objective of the next step:}

<objective of the next step>

\end{tcolorbox}
\

\textbf{Complete reasoning.} This prompt directly asks the LLM to complete the reasoning from the current reasoning state, corresponding to the simulation phase in MCTS. In memory-guided tree search, we perform completion after each expansion. Specifically, after applying a sub-instruction to generate a new reasoning step, we prompt the LLM to complete the remaining reasoning and produce the final solution.

\begin{tcolorbox}[
    breakable,
    colback=gray!10,
    colframe=gray!50,
    title=Complete reasoning,
    fonttitle=\bfseries,
    boxrule=0.5pt
]
\textbf{Instruction:} You are given a question and a partially completed reasoning chain. Continue the reasoning step by step, each step starts with 'Step $i$:', where $i$ is the step number, until the solution is complete. Do not restart from the beginning.

Present the final answer as (\textbackslash boxed\{X\}), where X is the correct answer.\\

\textbf{Question:}

<question>\\

\textbf{Reasoning so far:}

<reasoning so far>

\end{tcolorbox}
\

\textbf{Summarize sub-instructions.} This prompt asks the LLM to extract common reasoning patterns from multiple similar reasoning steps and summarize them into a sub-instruction. The resulting sub-instruction is then used to guide the generation of new reasoning steps when selected.

\begin{tcolorbox}[
    breakable,
    colback=gray!10,
    colframe=gray!50,
    title=Summarize sub-instructions,
    fonttitle=\bfseries,
    boxrule=0.5pt
]
\textbf{Instruction:} You are given several reasoning steps that are semantically similar.

Your task is to summarize them into a general instruction that captures the shared logic or operation behind these steps.\\

\textbf{Guidelines:} 

- Abstract away task-specific numbers or entities (replace them with variables or placeholders).

- Keep the instruction short and action-oriented (like a recipe step).

- If there are small differences across the steps, generalize to a pattern that covers them.

- Output only one generalized instruction.\\

\textbf{Reasoning Steps:}

<Reasoning Steps>\\

\textbf{Format:}

<general\_instruction>

[one-sentence generalized instruction here, using variables or placeholders if needed]

</general\_instruction>

\end{tcolorbox}
\

\textbf{Identify contextual conditions.} This prompt extracts the contextual conditions for a reasoning step from the question and preceding responses. The extracted conditions are used to build a codebook that identifies the most relevant context for applying the corresponding sub-instruction.

\begin{tcolorbox}[
    breakable,
    colback=gray!10,
    colframe=gray!50,
    title=Identify contextual conditions,
    fonttitle=\bfseries,
    boxrule=0.5pt
]
\textbf{Instruction:} Try to identify the INPUT CONDITIONS for the NEXT reasoning step only (not the whole solution) from Question and the Reasoning so far.

- Output the identified INPUT CONDITIONS only.\\

\textbf{Question:}

<question>\\

\textbf{Reasoning so far:}

<reasoning so far>\\

\textbf{Next reasoning step:}

<next reasoning step>

\end{tcolorbox}

\end{document}